# Lipi Gnani - A Versatile OCR for Documents in any Language Printed in Kannada Script

H. R. SHIVA KUMAR, Indian Institute of Science, India
A. G. RAMAKRISHNAN, Indian Institute of Science, India

A Kannada OCR, named Lipi Gnani, has been designed and developed from scratch, with the motivation of it being able to convert printed text or poetry in Kannada script, without any restriction on vocabulary. The training and test sets have been collected from over 35 books published between the period 1970 to 2002, and this includes books written in Halegannada and pages containing Sanskrit slokas written in Kannada script. The coverage of the OCR is nearly complete in the sense that it recognizes all the punctuation marks, special symbols, Indo-Arabic and Kannada numerals and also the interspersed English words. Several minor and major original contributions have been done in developing this OCR at the different processing stages such as binarization, line and character segmentation, recognition and Unicode mapping. This has created a Kannada OCR that performs as good as, and in some cases, better than the Google's Tesseract OCR, as shown by the results. To the knowledge of the authors, this is the maiden report of a complete Kannada OCR, handling all the issues involved. Currently, there is no dictionary based postprocessing, and the obtained results are due solely to the recognition process. Four benchmark test databases containing scanned pages from books in Kannada, Sanskrit, Konkani and Tulu languages, but all of them printed in Kannada script, have been created. The word level recognition accuracy of Lipi Gnani is 4% higher on the Kannada dataset than that of Google's Tesseract OCR, 8% higher on the datasets of Tulu and Sanskrit, and 25% higher on the Konkani dataset.

## 1 INTRODUCTION

Unlike the Roman script and Indian scripts such as Tamil, scripts such as Kannada and Telugu have a more complex arrangement of the letters in a text line. Figure 1a shows the image of a part of a printed Kannada page, illustrating the same. In Kannada, two or three consonants can combine together with a vowel and form a compound character or akshara. In such a case, the second and the third consonants become consonant conjuncts (called ottus) and appear at the bottom or right bottom of the first consonant. Interestingly, the vowel at the end of the consonant cluster modifies

Authors' addresses: H. R. Shiva Kumar, Indian Institute of Science, Department of Electrical Engineering, Bangalore, Karnataka, 560012, India, shivahr@gmail.com; A. G. Ramakrishnan, Indian Institute of Science, Department of Electrical Engineering, Bangalore, Karnataka, 560012, India, agr@iisc.ac.in.

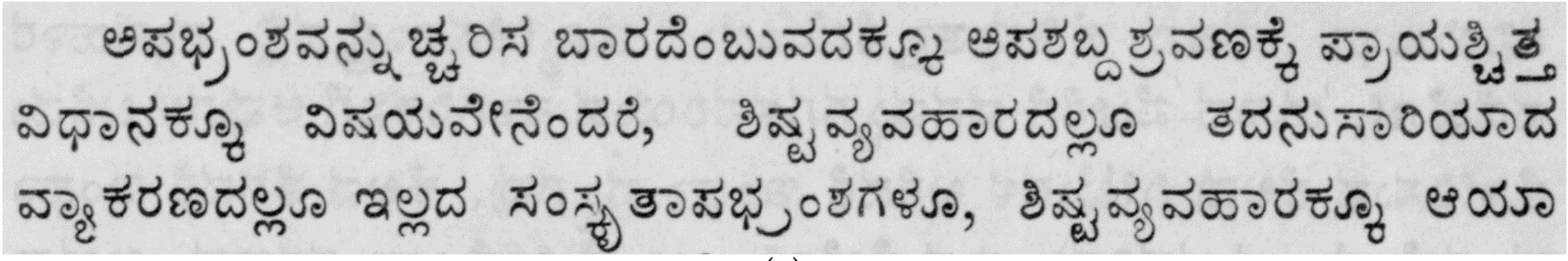

(a)

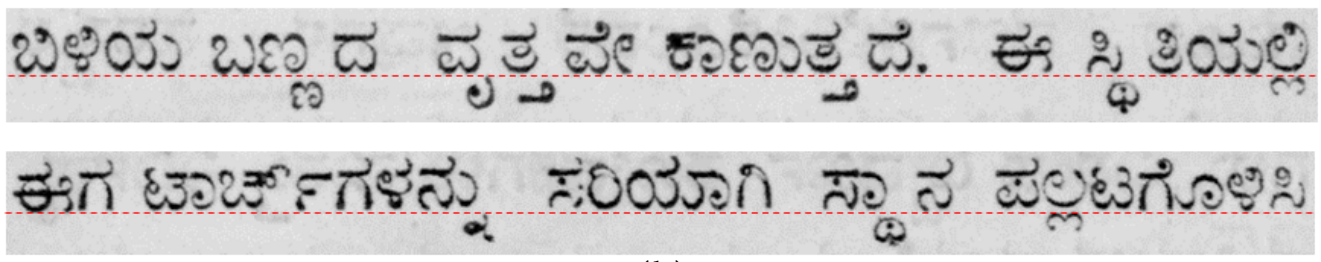

(b)

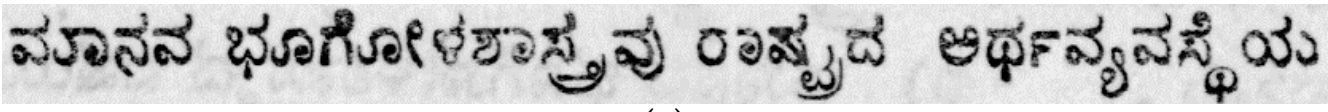

(c)

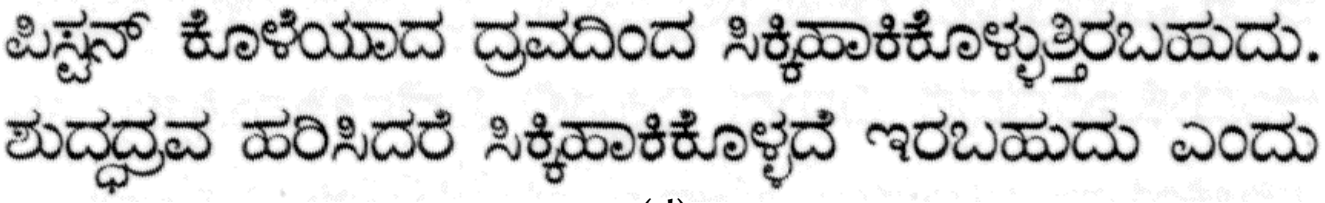

(d)

Fig. 1. Sample Kannada images, showing the consonant conjuncts (ottus) and their varying sizes and spatial positions with respect to the main text line. (a) Many ottus protruding up into the main text line. (b) Ottus form a separate horizontal line. (c) Two ottus stacked vertically one below the other. (d) Some of the ottus and base characters touching the other characters in the main text line and forming merged characters.

the shape of the first consonant present in the main text line. The symbols present in the main line are referred to as the base symbols. Some of the ottus (eg. ್ತ /ta_ottu/, ್ಗ /ga_ottu/, ್ದ /da_ottu/) are short and hence, when a text line contains only such short ottus, it is possible that the ottus form a separate horizontal line, not vertically overlapping with the main text line of base symbols, as shown by the example in Fig. 1b. On the other hand, some of the ottus have more height (eg. ್ಕ /ka_ottu/, ್ಯ /ya_ottu/, ್ರ /ra_ottu/) and when one or more such conjuncts are present, they vertically overlap with the bottom parts of the base characters and form a more complex line, as shown in the example in Fig. 1a. When the ್ರ /ra_ottu/ follows ್ತ /ta_ottu/ or ್ಟ /tta_ottu/, the two ottus are printed one below the other, forming a ottu complex, as shown in Fig. 1c. The situation becomes more complex, when a ottu or its ascender touches the base character above and forms a merged character, as shown in Fig. 1d.

Decades ago, the Halegannada (meaning old Kannada) had other consonants such as ಱ /rra/ (the stressed ರ /ra/) and ೞ /llla/ (the stressed ಳ /lla/), equivalents of which are still present in modern Tamil and Malayalam. So, the older poems and printed books contain these additional consonants and their vowel-modified forms. An OCR is most useful to digitize old, rather than new books. Thus, a good Kannada OCR needs to handle all of these different possibilities.

Further, many Kannadigas read the Sanskrit *slokas* (prayers) printed in Kannada script, because they are not very conversant with the Devanagari script, normally used for Sanskrit. In addition, the Dravidian language of Tulu [Wikipedia 2018b] and the Indo-Aryan language of Konkani [Wikipedia 2018a] do not have dedicated scripts of their own. Almost all of the Tulu works are printed in Kannada script, and Konkani works are printed in either Devanagari, Kannada or Malayalam scripts. We have developed an OCR that can cater to the digitization of all of these types of documents. Because of these requirements and motivation, we have not incorporated any post-processing based on any Kannada dictionary, since such a processing will work negatively for documents of non-Kannada origin.

Table 1. Kannada alphabets. The roman names given below each alphabet are as per the Kannada Unicode chart [Unicode 2018] except for the names Ru, am & ah used in place of the Unicode names vocalic_r, anusvara & visarga for saving space.

| ಅ | ಆ | ಇ | ಈ | ಉ | ಊ | ಋ | ಎ | ಏ | ಐ | ಒ | ಓ | ಔ | ಂ | ಃ |
|---|---|---|---|---|---|---|---|---|---|---|---|---|---|---|
| a | aa | i | ii | u | uu | Ru | e | ee | ai | o | oo | au | am | ah |

| | | | | |
|---|---|---|---|---|
| ಕ ka | ಖ kha | ಗ ga | ಘ gha | ಙ nga |
| ಚ ca | ಛ cha | ಜ ja | ಝ jha | ಞ nya |
| ಟ tta | ಠ ttha | ಡ dda | ಢ ddha | ಣ nna |
| ತ ta | ಥ tha | ದ da | ಧ dha | ನ na |
| ಪ pa | ಫ pha | ಬ ba | ಭ bha | ಮ ma |

| | | | | | | | | | | |
|---|---|---|---|---|---|---|---|---|---|---|
| ಯ ya | ರ ra | ಱ rra | ಲ la | ವ va | ಶ sha | ಷ ssa | ಸ sa | ಹ ha | ಳ lla | ೞ llla |

| ೦ | ೧ | ೨ | ೩ | ೪ | ೫ | ೬ | ೭ | ೮ | ೯ | ಽ | । | ॥ |
|---|---|---|---|---|---|---|---|---|---|---|---|---|
| zero | one | two | three | four | five | six | seven | eight | nine | avagraha | danda | double_danda |

Table 2. Illustration of formation of Kannada compound letters. Vowels in their dependent form (vowel sign) combined with the consonant ಕ್ /k/ forming various compound lettters of ಕ್ /k/.

| ಅ | ಆ | ಇ | ಈ | ಉ | ಊ | ಋ | ಎ | ಏ | ಐ | ಒ | ಓ | ಔ | ಂ | ಃ | | |
|---|---|---|---|---|---|---|---|---|---|---|---|---|---|---|---|---|
| | ◌ಾ | ◌ಿ | ◌ೀ | ◌ು | ◌ೂ | ◌ೃ | ◌ೆ | ◌ೇ | ◌ೈ | ◌ೊ | ◌ೋ | ◌ೌ | ◌ಂ | ◌ಃ | ◌್ | |
| ಕ | ಕಾ | ಕಿ | ಕೀ | ಕು | ಕೂ | ಕೃ | ಕೆ | ಕೇ | ಕೈ | ಕೊ | ಕೋ | ಕೌ | ಕಂ | ಕಃ | ಕ್ | ◌್ಕ |
| ka | kaa | ki | kii | ku | kuu | kRu | ke | kee | kai | ko | koo | kau | kam | kah | k | k_ottu |

## 1.1 Kannada Alphabets

Table 1 lists all the vowels (V), consonants (C) and numerals of Kannada script, as well as a few other symbols used as sentence-ending periods (danda or double-danda used in poetic text), vowel duration-extender (avagraha) and anusvara used for representing pure nasal consonants. Table 2 lists the modifiers associated with each of the vowels, which normally are connected to any of the consonants, to result in the corresponding C-V combination occurring as a single connected component. As an example, all the C-V combinations of the consonant ಕ /ka/ are listed. All these symbols form part of the classes recognized by our OCR. However, in many old, letter press printed books, these vowel modifiers have been printed separate from the consonants and hence, we have included the isolated vowel modifiers as additional recognition classes in our OCR. The complete list of recognition units are provided in the appendix.

## 1.2 Literature survey

The earliest known practical application of English OCR to postal address processing was by Srihari [Srihari 1993]. The first contribution in the development of OCR technology for Indic languages is by Chaudhuri and Pal for Bangla [Chaudhuri and Pal 1998]. This was shortly followed by Nagabhushan for Kannada [Nagabhushan and Pai 1999], Lehal for Gurmukhi [Lehal and Singh 2000], and Veena and Sinha for Devanagari [Bansal and Sinha 2000, 2002]. Negi and Chakravarthy reported Telugu OCR in 2001 [Negi et al. 2001]. Then, Sadhana had a special issue in 2002, where there were articles on the overview of document analysis [Kasturi et al. 2002], Kannada OCR [Ashwin and Sastry 2002], Odiya OCR [Chaudhuri et al. 2002], postprocessing of Gurmukhi OCR [Lehal and Singh 2002] and also identification of Tamil and Roman scripts at the level of words [Dhanya et al. 2002]. A Tamil OCR was also reported by two different groups [Aparna and Ramakrishnan 2002;

Kokku and Chakravarthy 2009]. BBN reported a Hindi OCR in 2005 [Natarajan et al. 2005]. A lot of work has been carried out by Manmatha related to historical documents [Rath and Manmatha 2007]. There was an update on most of the above works in the book edited by Govindaraju and Setlur [Govindaraju and Setlur 2009], and a Malayalam OCR was also reported [Neeba et al. 2009].

There are a number of reports in the literature on Kannada OCR [Ashwin and Sastry 2002; Nagabhushan and Pai 1999; Vijay Kumar and Ramakrishnan 2002, 2004]. However, most of these reports deal only with Kannada characters and none of them deal with the recognition of numerals and other symbols normally present in any printed document. Recently, Mathew et al. [Mathew et al. 2016] reported the use of recurrent neural networks with bidirectional long-short-term memory cells in the hidden layers and connectionist temporal classification in the output layer. The dataset used by the authors has 5000 printed pages for many languages, and for Kannada, the authors report the performance of 5.6% character error rate on 3500 pages. However, that database is not publicly available.

### 1.3 Contributions of the paper

- A realistic Kannada OCR has been designed and developed from scratch, which can actually recognize any old, printed page in Kannada, with a very good performance. It recognizes Indo-Arabic and Kannada numerals, Halegannada characters, as well as all the frequently used punctuation marks.
- A rich collection of benchmarking test datasets have been created, with annotated ground truth for Kannada, Konkani, Tulu and Sanskrit languages, all printed using Kannada script, and have been made openly available. Thus, researchers working on this topic can now compare their performance on standardized datasets.
- Results of our Lipi Gnani OCR on public datasets of Kannada, Tulu, Konkani and Sanskrit languages printed in Kannada script have been compared with Google's Tesseract OCR, as well as other available Kannada OCRs. These results will form reference results for future research on Kannada OCR.
- Of the Indian languages, Kannada has the most complex character set. This fact has also been acknowledged by Ray Smith in the OCR tutorial given by him in DAS 2016 [Smith 2016b]. We have handled the challenges posed by this character set in printed pages from really old books.
- Unicode generation in Kannada is involved, and occasionally one needs to deal with a sequence of four to five graphemes. In some cases, the sequence of Unicodes follows an order different from the order in which the corresponding symbols occur in the text. We have come out with systematic rules and flow-chart for Unicode generation from any sequence of recognized Kannada primitives.

## 2 THE DESIGN OF LIPI GNANI

Figure 2 gives the block diagram of the various steps performed in our Kannada OCR. The input image is first binarised. The skew angle is detected by Hough transform and corrected by bilinear interpolation. This gray image is once again binarized using Otsu thresholding. Then we extract the text blocks and segment them into lines, words, and characters/symbols. The symbols in our case are horizontally overlapping connected components as shown in Tables 4, 11 and 12. Symbol segmentation is followed by feature extraction, classification and Viterbi based selection of the best sequence of classifier outputs using symbol-level bigram probability. Finally, the Unicodes of the recognised symbols are concatenated in the correct order to generate the word level Unicode.

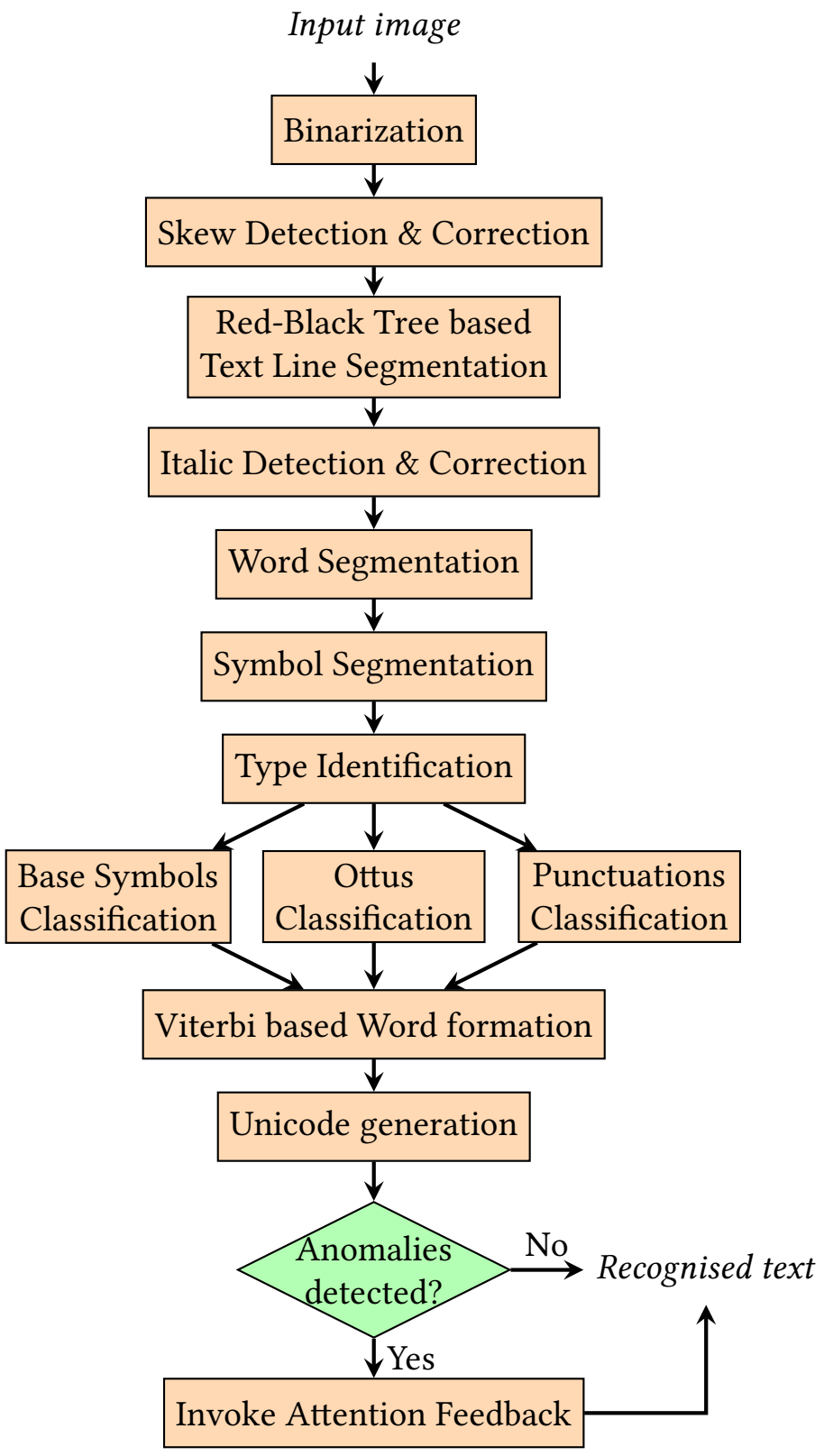

Fig. 2. Block Diagram of Lipi Gnani - the Kannada OCR.

### 2.1 Text Block Selection, Skew correction and Binarization

In our OCR, we have skipped the implementation of automatic text block extraction, but instead provide an intuitive GUI where users can draw rectangles and specify text blocks for a multi-column input page. If no text blocks are specified, then we assume the input image to contain a single text block.

To take care of any skew introduced during scanning, we detect the skew angle of the binarized image using Hough transform approach. If the detected skew angle $\theta_S$ is greater than an acceptable threshold ($0.5°$ in our OCR), then we correct the skew by rotating the image in the opposite direction (by $-\theta_S$ angle).

To make sure there is no loss of information during scanning, we recommend the input document to be scanned as a grey-level image at a minimum resolution of 300 dpi and the scanned image to be saved in uncompressed image format (like uncompressed TIFF). Since information that we need for text recognition is innately binary (text-ink present or not), we binarize the input image using an effective binarization algorithm to make sure that individual characters/symbols are not cut and consecutive symbols are not merged. Otsu's global thresholding algorithm [Otsu 1979] performs well for good quality documents and that is what we have implemented at the first level. However, in the case of degraded documents containing split or merged characters, we need locally adaptive binarization and this is invoked by the attention-feedback process indicated in Fig. 2. The details of this are given in the companion paper on attention feedback in OCR.

### 2.2 Line, Word Segmentation and Italic Correction

For text line segmentation, we use a bottom up approach based on the sizes and the relative positions of the connected components in the document page.

The line image is then sheared at different angles in small steps up to 20 degrees and the vertical projection profile (VPP) is taken at every step. The image corresponding to the VPP with maximum character/word gaps is considered as the italics corrected image and is used for word segmentation.

For word segmentation, we use the simple approach of classifying the gaps between symbols (zeros in vertical projection profile) into either inter-word gaps or intra-word gaps by using an adaptive threshold that varies with the height of the text line.

Kannada has a few consonants such as ಪ /pa/ and ಸ /sa/, which have 3 unconnected components one above the other. Therefore, in our segmentation, all the connected components which overlap in their VPP are clubbed together as a basic unit for recognition. Incidentally, this step takes care of any split in the character in the vertical direction.

### 2.3 Kannada Recognition Units and Symbol Segmentation

For symbol segmentation, we extract the connected components (CCs) and group CCs that have significant horizontal overlap above the base-line into one symbol. For example in the word ಸ್ವಂತ, the segmented symbols are ಸ, ್ವ, ಂ and ತ.

The set of symbols to be recognized was decided based on the commitment to design an OCR that works on both old and new Kannada books. Table 11 in the appendix lists the various distinct symbols used in Kannada script. Each of these symbols normally appears as a single connected component in the scanned image. The top row lists the independent vowels and visarga. The last row lists the Kannada numerals. The rest of the rows list the consonants with their various vowel combinations. The last but one column lists the pure consonants, which are consonants without any vowel attached to them. The last column lists the consonants in their *ottu* form. Ottus are symbols used for representing consonants when they appear as second or subsequent consonants in a consonant cluster. For example, in the word ಅಕ್ಕ /akka/, the symbol ್ಕ /ka_ottu/ is the ottu form of ಕ /ka/. Similarly, in the word ರಾಷ್ಟ್ರ /raassttra/, the symbols ್ಟ /tta_ottu/ and ್ರ /ra_ottu/ are the ottu forms of ಟ /tta/ and ರ /ra/, respectively.

The symbol ಂ shown in Table 3 can represent either anusvara (as in ಅಂ /am/, ಕಂ /kam/, ಚಂ /cam/) or numeral zero and needs to be disambiguated based on the word context. The symbol ೯ can represent either numeral nine or an alternate ottu form of ರ /ra/ called *arkaa-ottu* as in the word ಸೂರ್ಯ /suurya/. The symbol ೕ, henceforth referred to as dheergha, is used to represent the long (stressed) form of some of the dependent vowels as in ಕೀ /kii/, ಕೇ /kee/, ಕೋ /koo/. Similarly, the symbol ೃ represents the dependent form of vowel ಋ /vocalic_r/ as in ಕೃ /kRu/, and the symbol ೈ is the dependent form of vowel ಐ /ai/ as in ಕೈ /kai/. The symbol ೠ /vocalic_rr/ which was earlier taught as representing the longer form of vowel ಋ /vocalic_r/, is practically non-existent in actual text. Hence we have dropped it from our symbol set.

In old letterpress printed documents, like the example shown in Fig. 6c, the symbols representing some of the dependent vowels have been printed as separate characters after the base components. For example, ಕಾ /kaa/, ಕು /ku/, ಕೂ /kuu/, ಕೊ /ko/ and ಕೌ /kau/ are printed as ಕ ಾ , ಕ ು, ಕ ೂ, ಕೆ ೂ, ಕ ೌ, respectively. Similarly, the modifier symbol *halant* used for indicating any pure consonant has been printed separate from the base component (consonant part of CVs) as in ಕ ್. Also, the base components printed before ಾ, ೌ and ್ appear in a slightly cut form, as shown in the first column of Table 12 in the appendix.

Table 3. Kannada recognition units that need disambiguation based on previous unit and the dependent vowels that appear as separate symbols.

| Glyph | Possible Characters |
|---|---|
| ೦ | Anusvara or Kannada digit 0 |
| ೯ | Kannada digit 9 or arkaa-ottu (ರ ್) |

| ೀ /ii/ | |
|---|---|
| ಕಿ /ki/ | ಕೀ /kii/ |
| ಕೆ /ke/ | ಕೇ /kee/ |
| ಕೊ /ko/ | ಕೋ /koo/ |

| ೃ /Ru/ | |
|---|---|
| ಕ /ka/ | ಕೃ /kRu/ |

| ೈ /ai/ | |
|---|---|
| ಕೆ /ke/ | ಕೈ /kai/ |

Table 4. Additional Kannada recognition units formed by touching of multiple ottus.

| Glyph | Unicode Sequence | Example |
|---|---|---|
| ್ಕೃ /kRu_ottu/ | ್/-a/ ಕ/ka/ ೃ /Ru/ | ಸಂಸ್ಕೃತಿ /samskRuti/ |
| ್ಟ್ರ /ttra_ottu/ | ್/-a/ ಟ/tta/ ್/-a/ ರ/ra/ | ರಾಷ್ಟ್ರ /rassttra/ |
| ್ತ್ರ /tra_ottu/ | ್/-a/ ತ/ta/ ್/-a/ ರ/ra/ | ಶಾಸ್ತ್ರ /shastra/ |
| ್ಪ್ರ /pra_ottu/ | ್/-a/ ಪ/pa/ ್/-a/ ರ/ra/ | ಜಗತ್ಪ್ರಸಿದ್ಧ /jagatprasidda/ |
| ್ವ್ರ /vra_ottu/ | ್/-a/ ವ/va/ ್/-a/ ರ/ra/ | ದಿಗ್ವ್ರತ /digvrata/ |
| ್ರೈ /rai_ottu/ | ್/-a/ ರ/ra/ ೈ /ai/ | ಕ್ರೈಸ್ತ /kraista/ |

Table 5. Numerals and special symbols handled.

| | | | | | | | | | |
|---|---|---|---|---|---|---|---|---|---|
| 1 | 2 | 3 | 4 | 5 | 6 | 7 | 8 | 9 | ಽ |
| ? | ( | ) | [ | ] | : | ; | ! | / | \| |
| . | , | ‘ | ’ | ' | - | | | | |

Table 4 lists the additional recognition units necessiated by the touching of multiple ottus in an inseparable way, in some of the old books. In summary, all the symbols shown in Tables 11, 12, and 4 together form the complete set of 390 units that are recognised by our OCR.

### 2.4 Feature Extraction and Classification

The segmented components are resized to $32 \times 32$ size and discrete wavelet transform (DWT) features are extracted from them using Haar wavelet. In addition, horizontal and vertical autocorrelation-based features are also extracted and all the above three are concatenated to form a feature vector of dimension 3072. Support vector machine (SVM) classifier with linear kernel is used for classification. We have a total of 390 different classes and we have collected on the average, 200 training samples for each class. The SVM is trained with 5-fold cross-validation to choose the best value of the parameter C.

### 2.5 Handling of Ottus and Special Symbols

Ottu symbols are printed below the base symbols and most of them occur completely below the baseline. However, in the case of some ottus, a small portion extends up above the baseline, as shown in Figs. 3 and 1a. This relative position of the ottu symbols with respect to the text line is leveraged for recognizing them separately from the rest of the symbols. Once a connected component is identified as belonging to the ottu group, we recognize it by extracting features from its $32 \times 32$ normalized image and passing them to a SVM classifier trained on ottu symbols.

Table 5 lists the numerals and special symbols handled by our OCR. For symbols listed in the third row, namely the dot(.), comma(,), left single quotation(‘), right single quotation(’), apostrophe(') and hyphen(-), we use their position relative to the baseline, foreline, etc., their size relative to the line height and aspect ratio to recognize them directly, independent of the rest of the symbols. The numerals and special symbols listed in the first and second rows are recognized along with the rest of the base symbols. The symbol ಽ, called as avagraha, is used in Kannada script for representing elongated vowels in poetic text, mainly borrowed from Sanskrit.

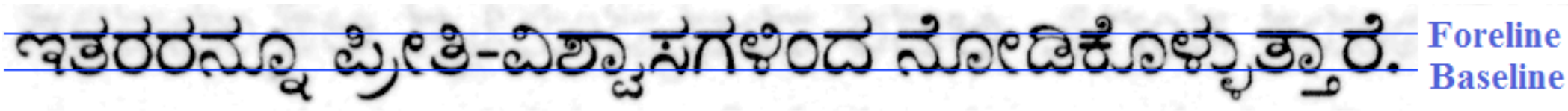

Fig. 3. Sample Kannada text line showing ottus and their position with respect to foreline and baseline.

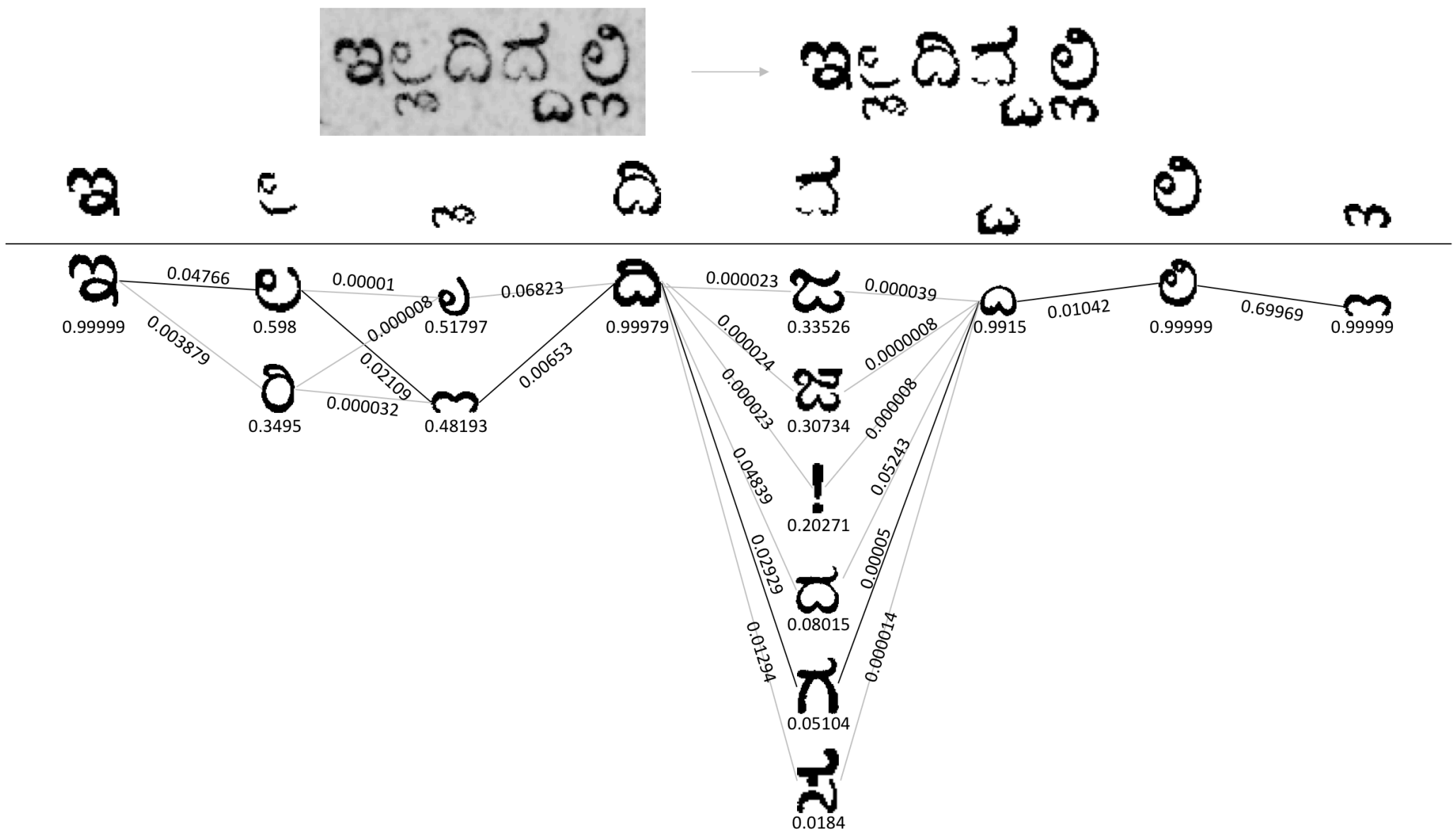

Fig. 4. Illustration of Viterbi based word formation, using a sample degraded word image.

### 2.6 Viterbi algorithm for word formation

In old books with poor print quality, portions of some of the characters are lost either directly in print or during binarization, as illustrated by the word image in Fig. 4. When such symbols are fed into the classifier, instead of getting a single recognized class with very high confidence level, we might get multiple labels, each with low confidence level. This is illustrated in Fig. 4, where each of the segmented components is shown above the horizontal line and the recognized classes, as output by the classifier, are shown below the horizontal line. The confidence level output by the classifier is shown below each of the recognized classes. In such cases, mere selection of the top recognized class may yield sub-optimal recognition results. For example, the incorrect Unicode sequence of ಇಲೃದಿಜ್ಡಲ್ಲಿ, for the word image shown in Fig. 4.

Leveraging symbol level bigram probabilities and the Viterbi algorithm [Forney 1973] for selecting the optimal sequence of recognized classes increases the chances of correct recognition, as shown in Fig. 4. Here, the symbol level bigram probabilities are shown on the edges joining the recognized classes and the optimal path selected by the Viterbi algorithm is shown using thick edges, yielding the correct Unicode output of ಇಲ್ಲದಿದ್ದಲ್ಲಿ for the word image shown in Fig. 4. For extracting the symbol level bigram probabilities, we have used a dump of Kannada Wikipedia text and our algorithm for transforming Kannada Unicode text to a sequence of Kannada OCR symbols.

Table 6. Unicodes corresponding to some sample recognition units of our Kannada OCR.

| | | | | | | | | | | |
|---|---|---|---|---|---|---|---|---|---|---|
| Symbol | ಕ /ka/ | ಕಾ /kaa/ | ಕಿ /ki/ | ಕು /ku/ | ಕೂ /kuu/ | ಕೆ /ke/ | ಕೊ /ko/ | ಕೌ /kau/ | ಕ್ /k/ | ್ಕ /k_ottu/ |
| Unicodes | ಕ | ಕ ಾ | ಕ ಿ | ಕ ು | ಕ ೂ | ಕ ೆ | ಕ ೊ | ಕ ೌ | ಕ ್ | ್ ಕ |

Table 7. The expected Unicode sequences for some sample Kannada words.

| Word | Unicode Sequence |
|---|---|
| ರಾಷ್ಟ್ರಪತಿ /raassttrapati/ | ರ/ra/ ಾ /aa/ ಷ/ssa/ ್/-a/ ಟ/tta/ ್/-a/ ರ/ra/ ಪ/pa/ ತ/ta/ ಿ /i/ |
| ಸಂಸ್ಕೃತಿ /samskRuti/ | ಸ/sa/ ಂ/am/ ಸ/sa/ ್/-a/ ಕ/ka/ ೃ /Ru/ ತ /ta/ ಿ /i/ |
| ಕ್ರೈಸ್ತ /kraista/ | ಕ/ka/ ್/-a/ ರ/ra/ ೈ /ai/ ಸ/sa/ ್/-a/ ತ/ta/ |
| ಸೂರ್ಯ /suurya/ | ಸ/sa/ ೂ /uu/ ರ/ra/ ್/-a/ ಯ/ya/ |

## 2.7 Kannada Unicode generation module

After symbol segmentation and classification, the recognised labels need to be put together in the correct order so as to generate a valid Unicode sequence [Unicode 2018] at the word level. Table 6 gives the Unicode sequences for a few recognition units used in our OCR. Table 7 lists the expected Unicode sequences for some sample Kannada words. For the word ರಾಷ್ಟ್ರಪತಿ /raassttrapati/, the sequence of segmented and recognised symbols are ರಾ /raa/, ಷ /ssa/, ್ಟ /tta_ottu/, ್ರ /r_ottu/, ಪ /pa/ and ತಿ /ti/. Putting together the Unicodes associated with these symbols in the same order, we would get ರ ಾ ಷ ್ ಟ ್ ರ ಪ ತ ಿ, which matches with the expected Unicode for that word.

However, for the word ಕ್ರೈಸ್ತ /kraista/, the sequence of segmented and recognised symbols are ಕೆ /ke/, ್ರ /r_ottu/, ೖ /vowel_sign_ai/, ಸ /sa/ and ್ತ /t_ottu/. Putting their associated Unicodes in the same order would give us ಕ ೆ ್ ರ ೈ ಸ ್ ತ, but the correct Unicode for that word is ಕ ್ ರ ೈ ಸ ್ ತ. Similarly, for the word ಸೂರ್ಯ /suurya/, the sequence of segmented and recognised symbols are ಸೂ /suu/, ಯ /ya/ and ರ್ /arkaaottu/. Putting their associated Unicodes in the same order would give us ಸ ೂ ಯ ರ ್, but the correct Unicode for that word is ಸ ೂ ರ ್ ಯ.

To generate the correct Unicode at the word level, we use the logic illustrated by the flowcharts of Fig. 5. From the list of recognised symbols, we first combine part characters as per Table 12. For example, if the recognised symbols have the sequence ಕ and ಾ, then we replace it by ಕಾ. Similarly, ನ and ು is replaced by ಸು, and, ವ and ಾ is replaced by ಮಾ. The next step, as shown in Fig. 5a, is to group the sequence of recognised symbols into aksharas. Symbols that represent independent vowels, consonants with/without dependent vowels, numerals and special symbols mark the beginning of aksharas. Symbols ೕ, ೖ, ೈ and ರ್ and ottu symbols get added to the previous akshara. In order to keep the logic simple, we consider anusvara and visarga as separate aksharas. Then, we generate the Unicode for each akshara as shown in Fig. 5b, and concatenate them to get the word level Unicode. The idea for generating the correct Unicode at the akshara level is to concatenate the Unicodes of all the symbols in it with the following exceptions:

- If the akshara has ರ್, then insert it at the beginning of akshara. For example, in the word ಸೂರ್ಯ, for the second akshara ರ್ಯ, we now get the correct Unicode ರ ್ ಯ.
- If the akshara begins with the symbol of a consonant-vowel combination, then we take the dependent vowel out and insert it at the end of the akshara. If the akshara has other vowel modifier symbols in it (such as ೕ, ೖ or ೈ), then we combine it with the first symbol's dependent vowel as per Fig. 5c, and insert the combined vowel modifier at the end of the akshara. For example, in the word ಕ್ರೈಸ್ತ, for the first akshara ಕ್ರೈ, the first symbols's vowel modifier ಿ combines with the vowel modifier symbol ೖ to give the Unicode ೈ, which is then inserted at the end of the akshara, to get the correct Unicode sequence ಕ ್ ರ ೈ.

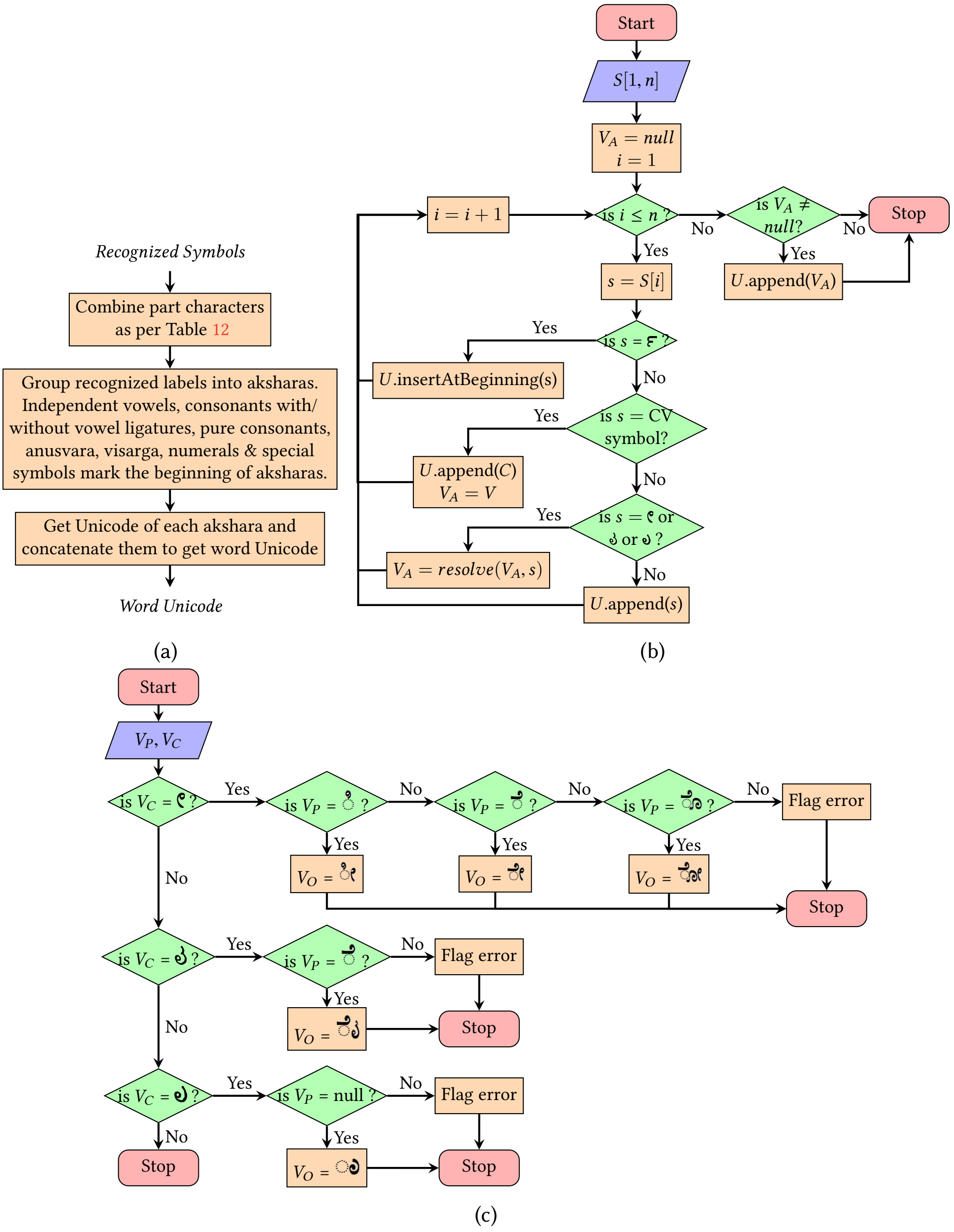

Fig. 5. (a) Block diagram for getting the word Unicode from the recognized labels. (b) Flowchart to generate Unicode for any akshara. The input is the list of recognised labels $S$. The output is the list of Unicodes $U$. $V_A$ is used for storing the vowel to be inserted at the end of the akshara. (c) Flowchart for resolving the vowel sign within the akshara as per Table 3. $V_P$ and $V_C$ are the labels of the previous and current vowel modifiers. $V_O$ is the resolved label of the output vowel modifier.

### 2.8 Attention Feedback in OCR

The performance of the OCR is further improved, based on feedback of anomalies detected from the recognized text at the output of the OCR, to various preprocessing stages, such as binarization, segmentation and also to the recognition stage, for splitting the merged characters, merging the split characters and recognizing the interspersed English words. A complete description of all these modules requires a very detailed discussion, and hence, the details are given in the accompanying paper on the role and effectiveness of attention-feedback in our OCR.

## 3 PERFORMANCE EVALUATION - OPEN DATASETS AND METRICS

### 3.1 Benchmark dataset of Kannada printed pages

The Kannada dataset contains 241 images, carefully chosen to have various kinds of recognition challenges. Figure 6 gives some examples from this benchmark dataset. Some of the pages have italics and bold characters (see Fig. 6a); some of them have Halegannada poems and text (see Fig. 6b); others are letterpress-printed pages, where the vowel modifiers appear as separate symbols and do not touch the consonants they go with (see Fig. 6c). Some pages have interspersed English words (see Fig. 6d); still others have tables with a lot of numeric data (see Fig. 6e). In addition, there are old pages containing either a lot of broken characters or many words with two or more characters merged into a single connected component. This rigorous dataset will be made public at the time of publishing of this paper.

### 3.2 Multilingual datasets printed in Kannada script

The Konkani dataset [MILE 2018a] contains 40 pages, scanned from different books. The Tulu [MILE 2018c] and the Sanskrit [MILE 2018b] datasets have 10 and 60 pages, respectively. All of these pages have been scanned as gray level images at 300 dpi resolution, and have Unicode ground truth, together with line, word and character coordinate information. These datasets with text coming from other languages are good testing ground for the raw recognition accuracy of the OCR, since many of the words in them may not be present in the dictionary of a typical Kannada OCR.

### 3.3 Computation of Recognition Accuracy

Let $N$ be the ground truth size (number of Unicodes for Unicode level accuracy OR number of words for word level accuracy) & $M$, the OCR output size. We use Levenshtein distance to compute the number of substitutions $S$, insertions $I$ and deletions $D$, both at the levels of the character (Unicode) and the word.

The total error measured and the recognition accuracy are defined as:

$$Error = S + I + D \quad (1)$$

$$Accuracy = (N - Error)/N \quad (2)$$

## 4 RESULTS AND DISCUSSION

### 4.1 Tesseract version compared

We compare our recognition results on all the test sets with those of the Google's Tesseract OCR. Tesseract was originally developed at Hewlett Packard (HP) between 1985 and 1994, and was open sourced by HP in 2005. Since 2006, it is being developed by Google. Thus, this OCR development has

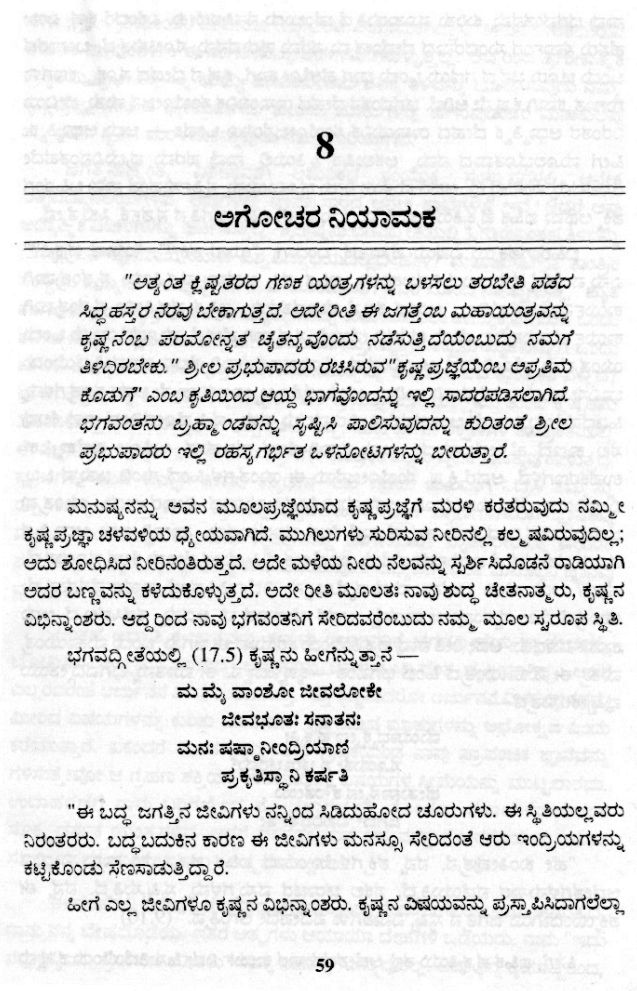

(a) Kannada page with italic and bold text.

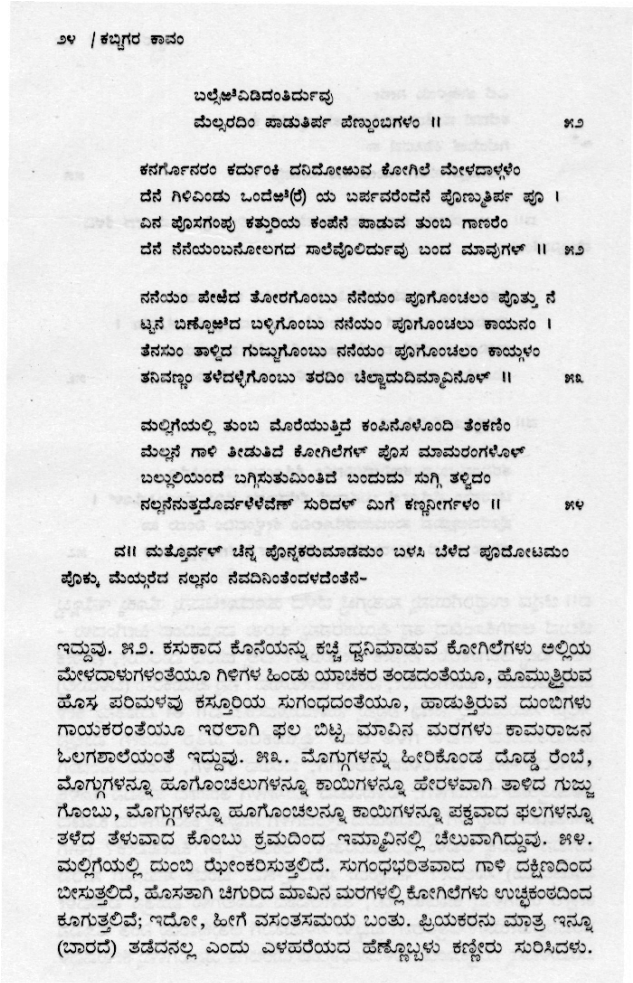

(b) Kannada page with halegannada letters and text.

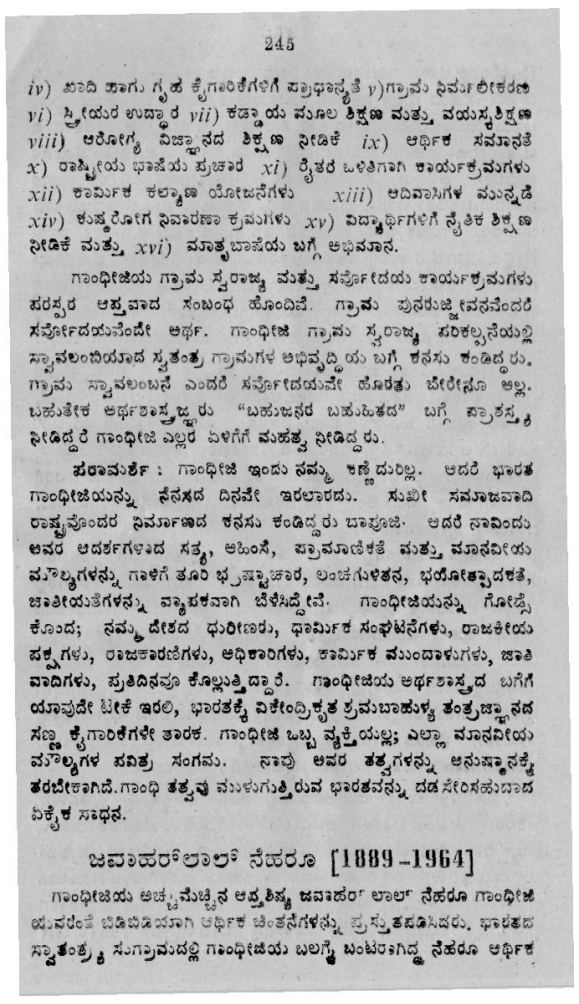

(c) Letterpress printed Kannada page: non-joining consonant modifiers

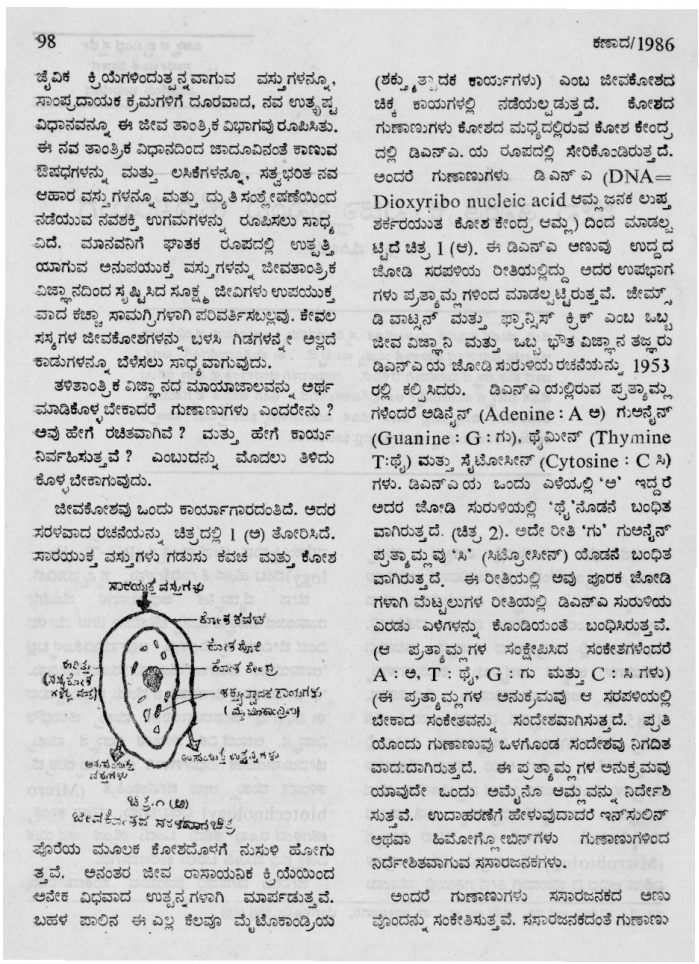

(d) Two column Kannada page with interspersed English words

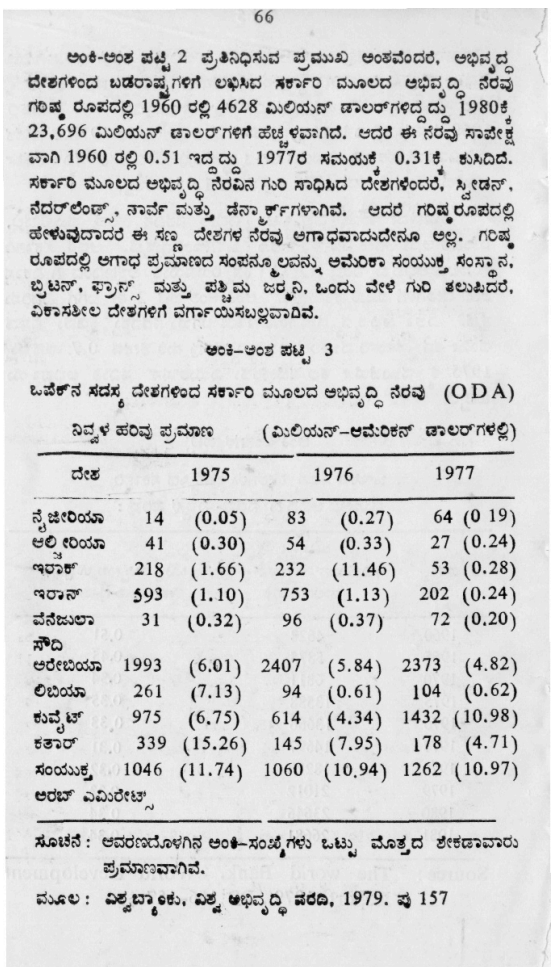

(e) Kannada page with table and numeric text

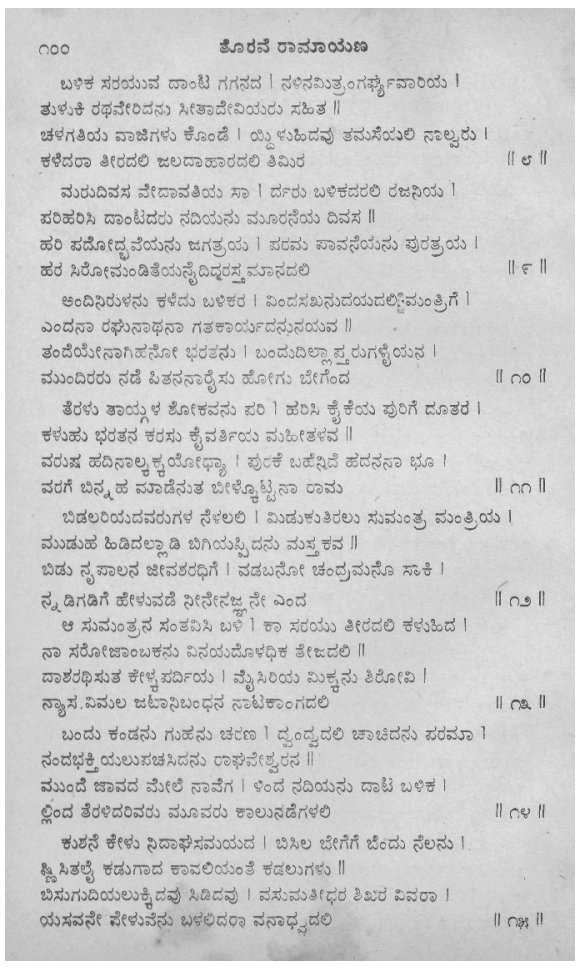

(f) Kannada page with old poetic text

Fig. 6. Sample images from Kannada benchmarking dataset

a history of over 30 years, backed by two multinational companies. In DAS-2016 Tutorial [Smith 2016a], Tesseract's T-LSTM engine (available in v4.00) was reported to have the best Unicode and word-level recognition accuracies for Kannada language. Hence, we tested on the latest version (v4.0.0) of Tesseract with command line option –oem 1 [Tesseract 2018].

The focus of our testing is on Unicode and word level recognition accuracies. Hence, to make sure that any possible mistakes in the layout analysis (like text/image block segmentation) does not impact the recognition accuracies, we manually pass the boundary coordinates of the text region of the test images to Tesseract through uzn [White 2014] files.

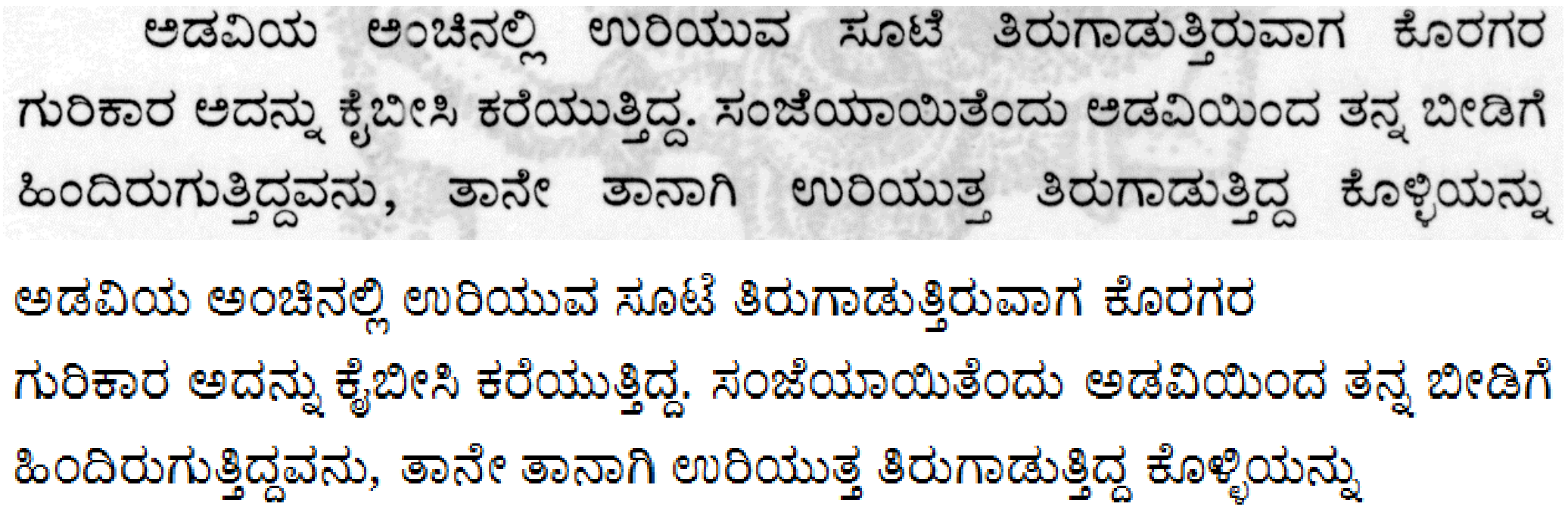

Fig. 7. A segment of a Kannada document, where the picture from the other side of the paper is seen through in the scanned image) and the text recognized by Lipi Gnani. UA = 99.09%, WA = 95.12% for the entire page.

Table 8. Unicode and word recognition accuracies of Lipi Gnani and Tesseract OCRs on Kannada dataset of 241 images. UA, WA: Unicode and word recognition accuracy. N, M: # of Unicodes in the ground-truth and the OCR output, respectively. S, I and D: # of Unicode substitutions, insertions and deletions w.r.t. the ground truth. $N_W$, $M_W$: # of words in the ground-truth and the OCR output, respectively. $S_W$, $I_W$, $D_W$: # of word substitutions, insertions and deletions w.r.t. the groundtruth. N = 3,97,371; $N_W$ = 47,464.

| | UA (%) | M | S | I | D | $M_w$ | $S_w$ | $I_w$ | $D_w$ | WA(%) |
|---|---|---|---|---|---|---|---|---|---|---|
| Lipi Gnani | 95.63 | 3,96,529 | 9,588 | 3,470 | 4,312 | 47,556 | 8,718 | 759 | 667 | 78.63 |
| Tesseract | 94.82 | 4,01,791 | 10,190 | 7,413 | 2,999 | 48,370 | 10,459 | 1338 | 432 | 74.24 |

### 4.2 Results on Kannada benchmarking dataset

Figures 7 to 10 display the results obtained from our OCR for some sample image segments from the test database. Figure 7 shows a small part of the scanned image of a Kannada printed page, where the picture on the other side of the paper is clearly seen through. Our OCR recognizes this image perfectly, with no errors (see bottom portion of Fig. 7). Figure 8 shows a portion of the scanned image of a Kannada printed page with half of the text bold and the rest normal. The text recognized by Lipi Gnani OCR has only two errors (see bottom portion of Fig. 8). Figure 9 shows a segment of the scanned image of a Kannada printed page with all italic text. The text recognized by our OCR has very few errors (see bottom portion of Fig. 9). Finally, Fig. 10 shows a small section of the scanned image of a highly degraded Kannada printed page. This is letter press printed, and hence, all the vowel modifiers appear clearly separated from their corresponding consonants. In spite of the degradations present, the text recognized by our OCR has just three errors (see bottom portion of Fig. 10).

Figure 11 compares the performances of Lipi Gnani and Tesseract OCRs on the benchmark Kannada dataset. Plotted in the two subfigures are the Unicode and word error rates obtained by the two OCRs on each of the pages. The UA and hence, the WA obtained by our OCR is lower in nearly 200 of the 241 images in the database. Table 8 compares the average accuracies of the two OCRs on the whole dataset. Lipi Gnani performs marginally better, with an improvement of 0.81% in UA and 4.39% in WA. Table 9 compares the processing times of the two OCRs and we can see that Lipi Gnani takes only 1.5 seconds per page, as against Tesseract's 5 seconds per page.

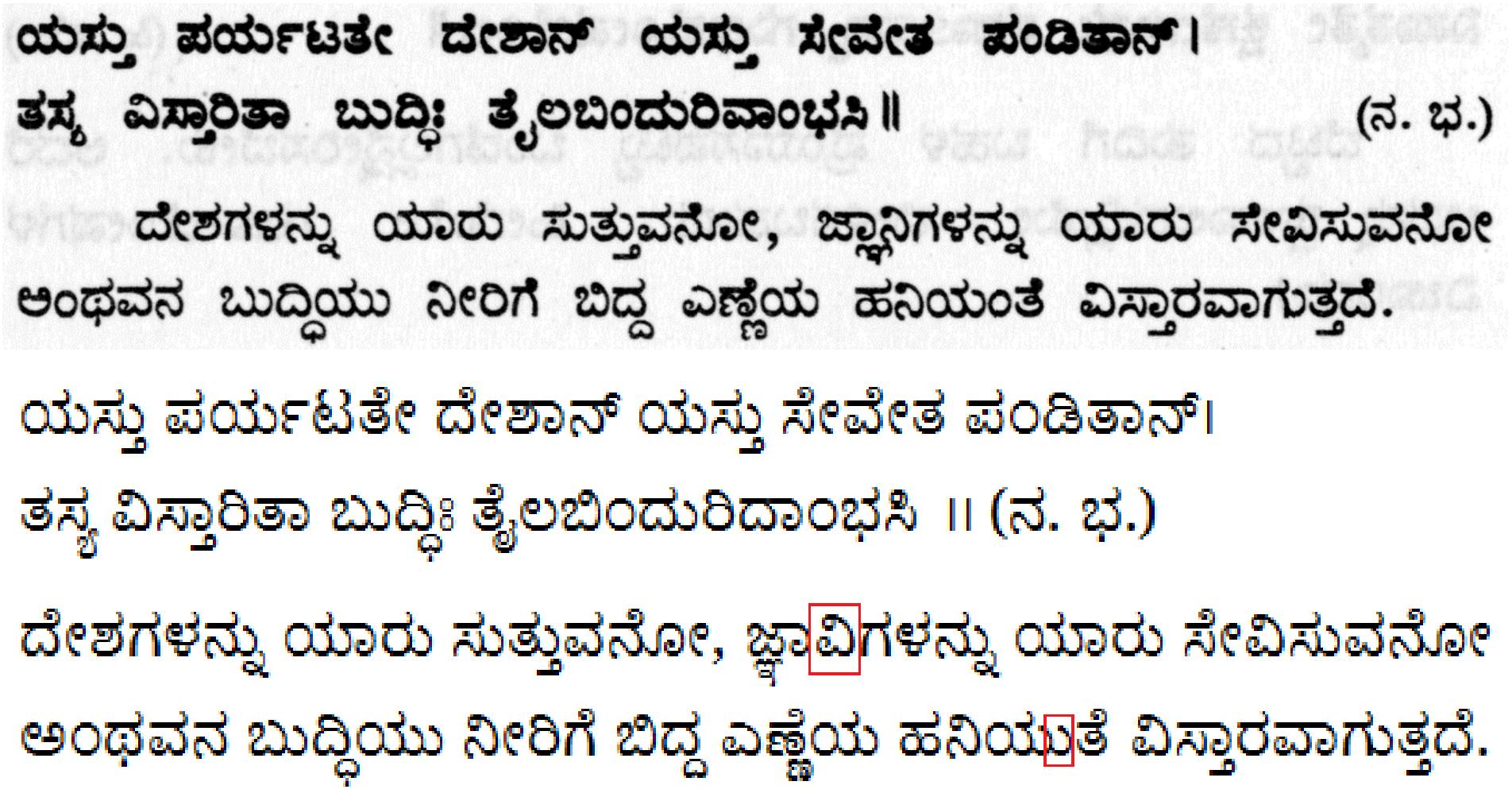

Fig. 8. A small part of the scanned image of a Kannada document with both bold and normal text), and the text recognized by Lipi Gnani. UA = 96.08%, WA = 78.91% for the full page. The two red bounding boxes indicate wrongly recognized characters.

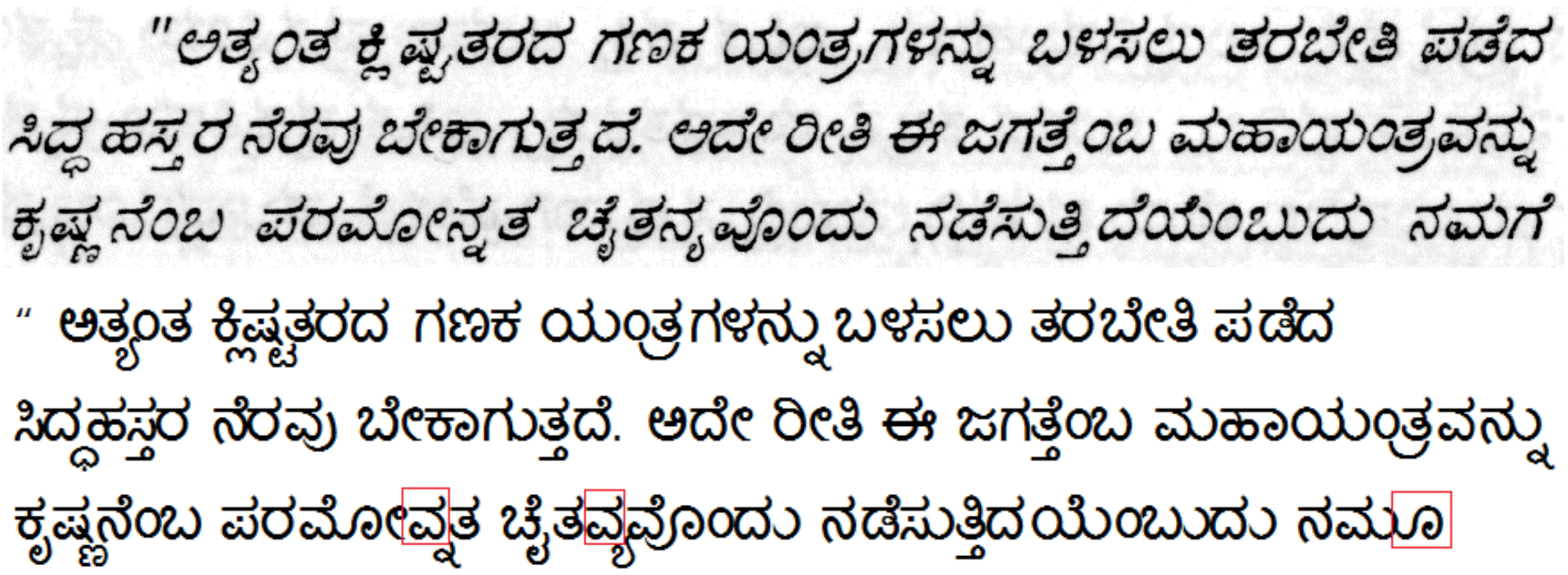

Fig. 9. A portion of a Kannada document image with italic text and the output text from the Lipi Gnani OCR, with very few errors. UA = 98.64%, WA = 89.47% for the entire page.

Table 9. Processing times of Lipi Gnani and Tesseract OCRs on Kannada benchmarking dataset of 241 images, run on a Windows 7 desktop having Intel i7 Octacore CPU @3.4 GHz with 16 GB RAM.

| | Total time (241 images) | Average time per page |
|---|---|---|
| Lipi Gnani | 6 mins | 1.5 secs |
| Tesseract | 20 mins | 5 secs |

### 4.3 Results on Konkani, Tulu, and Sanskrit benchmarking datasets

Figures 12 to 15 display the recognition results from our OCR for segments from one or two sample pages from each of the non-Kannada language text printed in Kannada. Along with the recognized text, each figure also shows the overall UA and WA achieved on the whole page, from which the segment being displayed arises. A small part of the scanned image of a Konkani document is shown at the top of Fig. 12. Also shown is the text output from our OCR, at the bottom part of the figure. Figure 13 shows a portion of the scanned image of another Konkani printed page. The text recognized by Lipi Gnani OCR, shown on the right, is nearly perfect. Figure 14 shows another example from a Sanskrit printed page. The text recognized by our OCR has only one wrong character (see the bottom of Fig. 14). Another Sanskrit example is shown in Fig. 15. Again, there is a single error in the text recognized by our OCR (see the right hand side of Fig. 15).

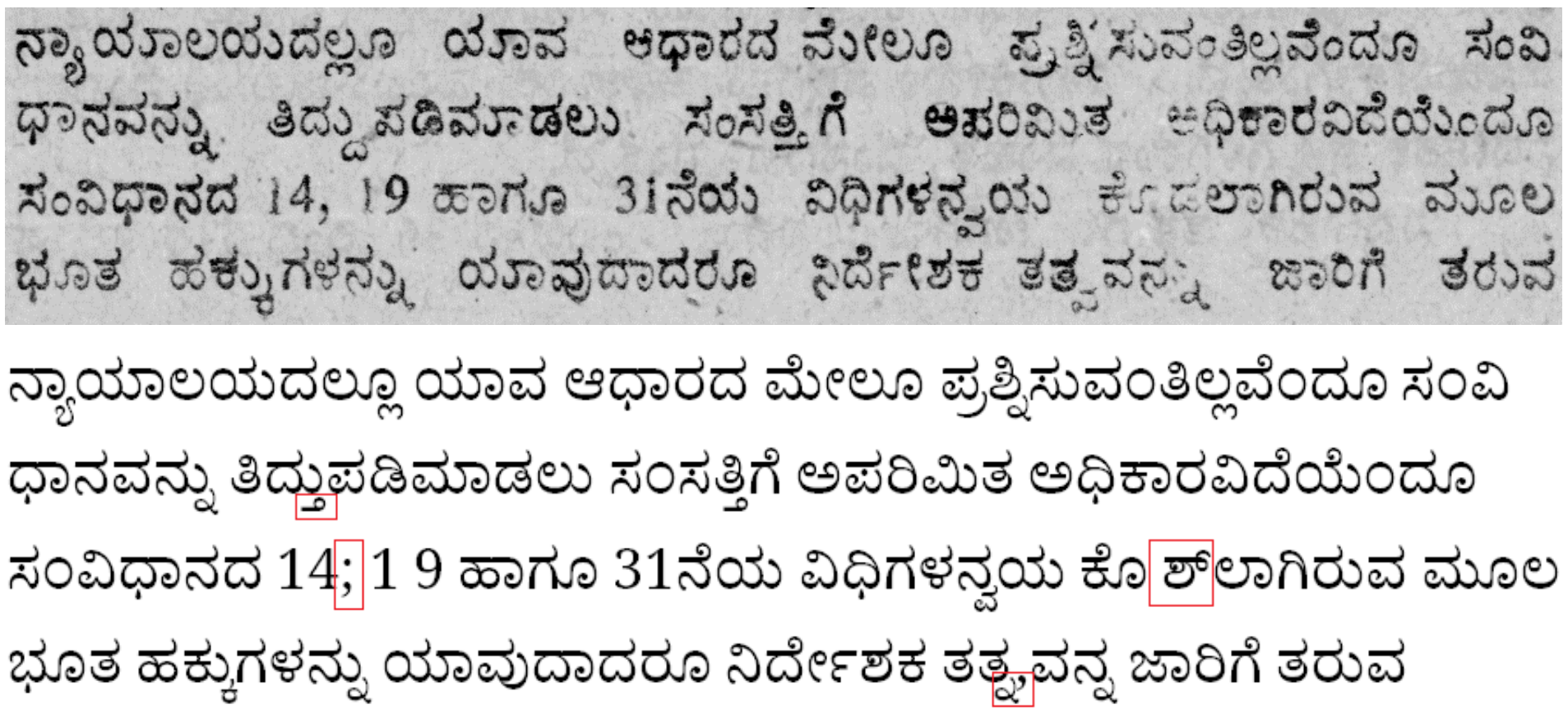

Fig. 10. A segment of an old, letter press printed, Kannada document page with severe localized degradations, and the text output by Lipi Gnani with negligible recognition errors. UA = 95.35% and WA = 76.96% for the complete page.

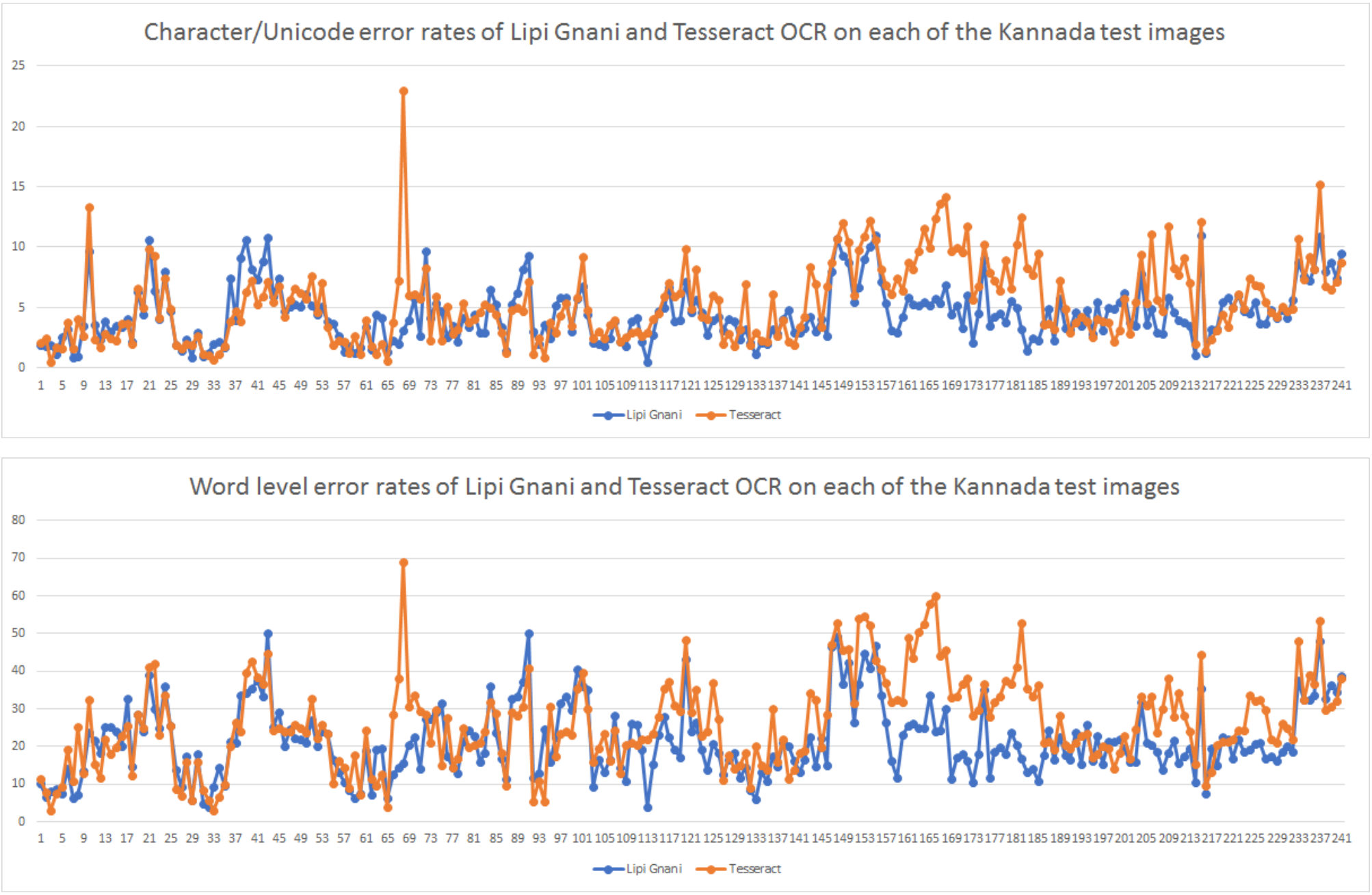

Fig. 11. Comparison of Unicode and word level error rates of Lipi Gnani and Tesseract OCR on each page of Kannada benchmarking dataset.

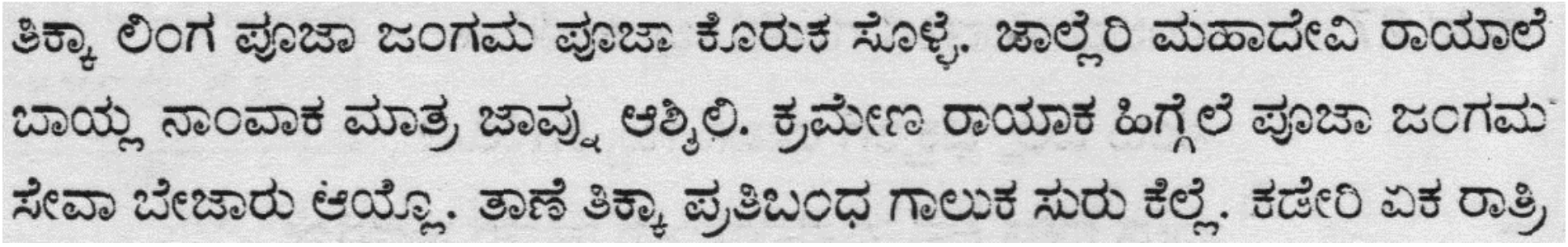
ತಿಕ್ಕಾ ಲಿಂಗ ಪೂಜಾ ಜಂಗಮ ಪೂಜಾ ಕೊರುಕ ಸೊಳ್ಳೈ. ಜಾಲ್ಲೆರಿ ಮಹಾದೇವಿ ರಾಯಾಲೆ
ಬಾಯ್ಲ ನಾಂವಾಕ ಮಾತ್ರ ಜಾವ್ನ ಆಶ್ಶಿಲಿ. ಕ್ರಮೇಣ ರಾಯಾಕ ಹಿಗ್ಗೆಲೆ ಪೂಜಾ ಜಂಗಮ
ಸೇವಾ ಬೇಜಾರು ಆಯ್ಲೊ. ತಾಣೆ ತಿಕ್ಕಾ ಪ್ರತಿಬಂಧ ಗಾಲುಕ ಸುರು ಕೆಲ್ಲೆ. ಕಡೇರಿ ಏಕ ರಾತ್ರಿ

ತಿಕ್ಕಾ ಲಿಂಗ ಪೂಜಾ ಜಂಗಮ ಪಆಜಾ ಕೊರುಕ ಸೊಳ್ಳೈ. ಜಾಲ್ಲೆರಿ ಮಹಾದೇವಿ ರಾಯಾಲೆ
ಬಾಯ್ಲ ನಾಂವಾಕ ಮಾತ್ರ ಜಾವ್ನ ಆಶ್ಬಿಲಿ. ಕ್ರಮೇಣ ರಾಯಾಕ ಹಿಗ್ಗೆಲೆ ಪೂಜಾ ಜಂಗಮ
ಸೇವಾ ಬೇಜಾರು ಕಯ್ಲೊ. ತಾಣೆ ತಿಕ್ಕಾ ಪ್ರತಿಬಂದ ಗಾಲುಕ ಸುರು ಕೆಲ್ಲೆ. ಕಡೇರಿ ಏಕ ರಾತ್ರಿ

Fig. 12. A segment from the scanned image of a Konkani document and the corresponding text recognized by Lipi Gnani. UA = 94.38% and WA = 79.15% for the full page.

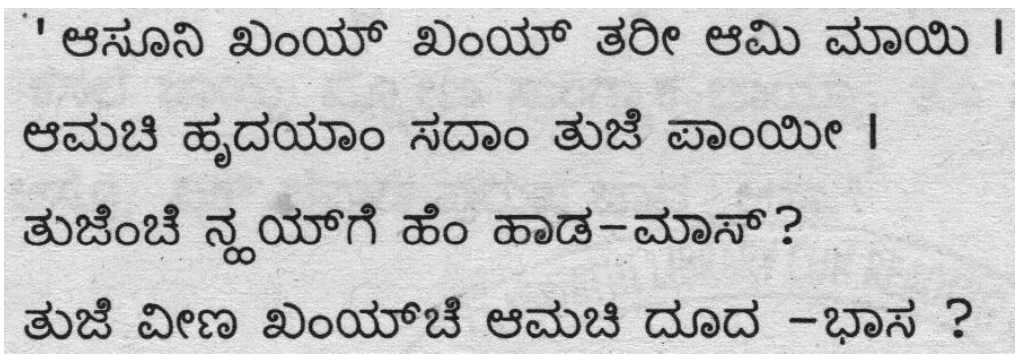
'ಆಸೂನಿ ಖಂಯ್ ಖಂಯ್ ತರೀ ಆಮಿ ಮಾಯಿ ।
ಆಮಚಿ ಹೃದಯಾಂ ಸದಾಂ ತುಜೆ ಪಾಂಯೀ ।
ತುಜೆಂಚೆ ನ್ವಯ್‌ಗೆ ಹೆಂ ಹಾಡ–ಮಾಸ್?
ತುಜೆ ವೀಣ ಖಂಯ್‌ಚೆ ಆಮಚಿ ದೂದ –ಭಾಸ ?

’ಆಸೂನಿ ಖಂಯ್ ಖಂಯ್ ತರೀ ಆಮಿ ಮಾಯಿ ।
ಆಮಚಿ ಹೃದಯಾಂ ಸದಾಂ ತುಜೆ ಪಾಂಯೀ ।
ತುಜೆಂಚೆ ನ್ವಯ್‌ಗೆ ಹೆಂ ಹಾಡ-ಮಾಸ್ ?
ತುಜೆ ವೀಣ ಖಂಯ್‌ಚೆ ಆಮಚಿ ದೂದ -ಭಾಸ ?

Fig. 13. A part of Konkani document page containing poetic text from another book, and the recognized text from our OCR. UA = 96.11% and WA = 79.26% for the entire page.

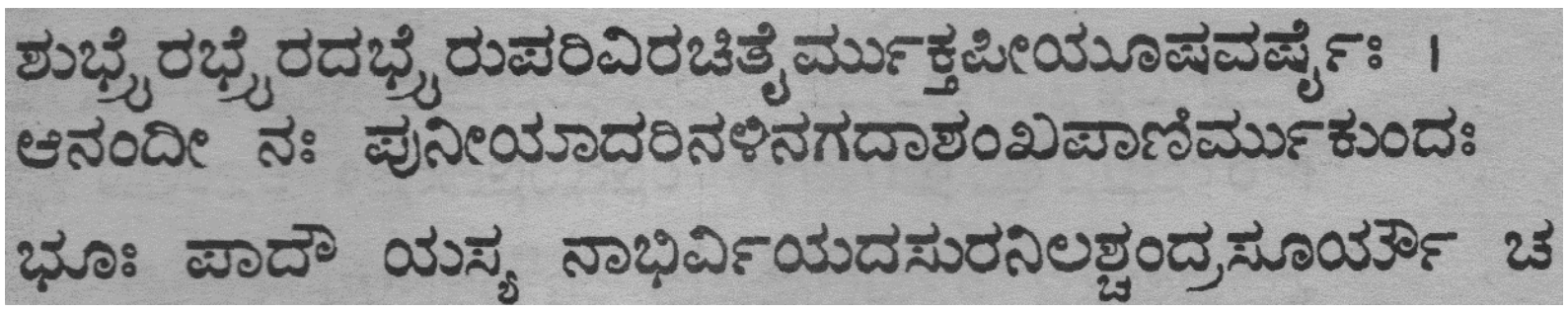
ಶುಭ್ರೈರಭ್ರೈರದಭ್ರೈರುಪರಿವಿರಚಿತೈರ್ಮುಕ್ತಪೀಯೂಷವರ್ಷೈಃ ।
ಆನಂದೀ ನಃ ಪುನೀಯಾದರಿನಳಿನಗದಾಶಂಖಪಾಣಿರ್ಮುಕುಂದಃ
ಭೂಃ ಪಾದೌ ಯಸ್ಯ ನಾಭಿರ್ವಿಯದಸುರನಿಲಶ್ಚಂದ್ರಸೂರ್ಯೌ ಚ

ಶುಭ್ರೈರಭ್ರೈರದಭ್ರೈರುಪರಿವಿರಚಿತೈರ್ಮುಕ್ತಪೀಯೂಷವರ್ಷೈಃ ।
ಆನಂದೀ ನಃ ಪುಮಿಯಾದರಿನಳಿನಗದಾಶಂಖಪಾಣಿರ್ಮುಕುಂದಃ ।।
ಭೂಃ ಪಾದೌ ಯಸ್ಯ ನಾಭಿರ್ವಿಯದಸುರನಿಲಶ್ಚಂದ್ರಸೂರ್ಯೌ ಚ

Fig. 14. A snippet from Sanskrit document and recognized text from Lipi Gnani. UA = 92.13% and WA = 71.88% for the full page.

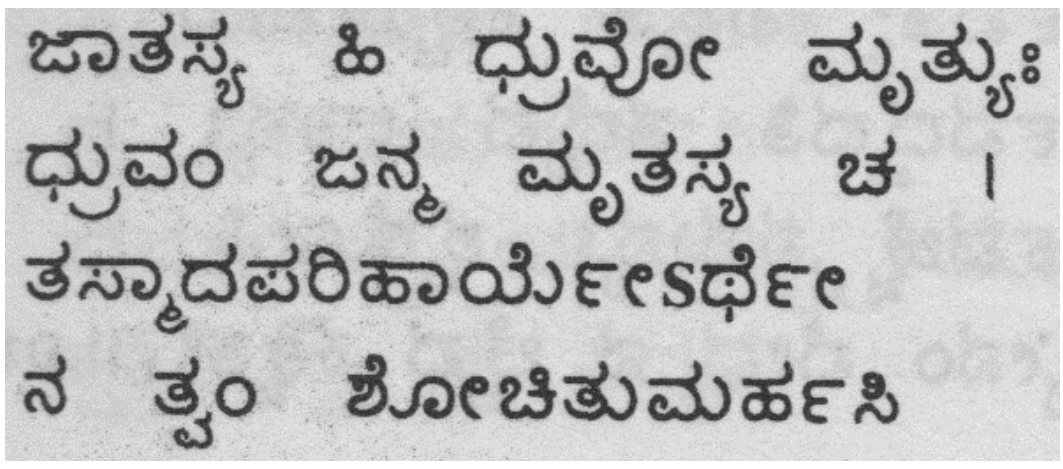
ಜಾತಸ್ಯ ಹಿ ಧ್ರುವೋ ಮೃತ್ಯುಃ
ಧ್ರುವಂ ಜನ್ಮ ಮೃತಸ್ಯ ಚ ।
ತಸ್ಮಾದಪರಿಹಾರ್ಯೇಽರ್ಥೇ
ನ ತ್ವಂ ಶೋಚಿತುಮರ್ಹಸಿ

ಜಾತಸ್ಯ ಹಿ ಧ್ರುವೋ ಮೃತ್ಯುಃ
ಧ್ರುವಂ ಜನ್ಬಮೃತಸ್ಯ ಚ ।
ತಸ್ಮಾದಪರಿಹಾರ್ಯೇಽರ್ಥೇ
ನ ತ್ವಂ ಶೋಚಿತುಮರ್ಹಸಿ

Fig. 15. A snippet from Sanskrit document and recognized text from Lipi Gnani. UA = 93.6% and WA = 67.35% for the full page.

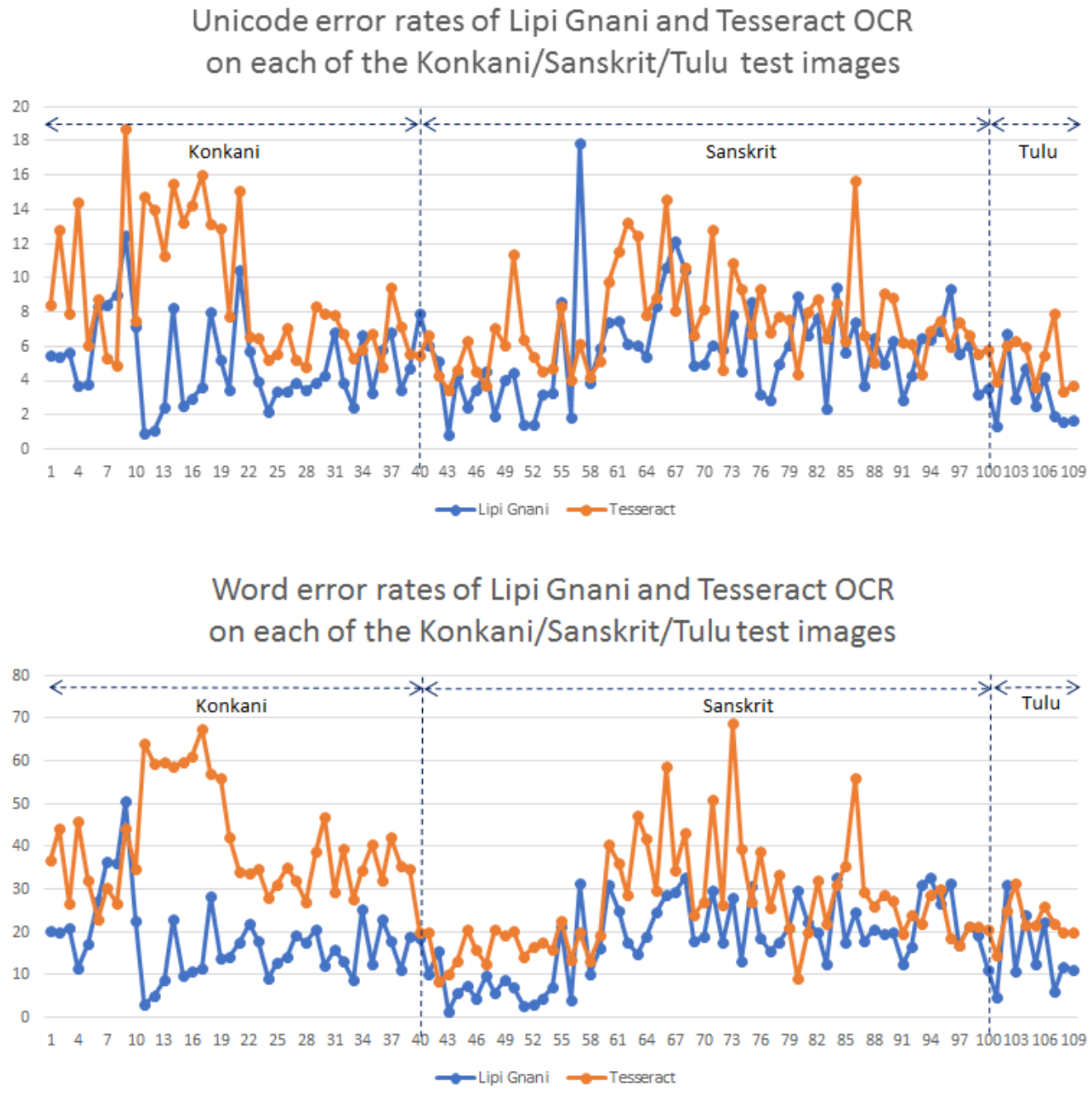

Fig. 16. Comparison of Unicode and word level error rates of Lipi Gnani and Tesseract OCR on each page of Konkani, Tulu and Sanskrit benchmark datasets.

Figure 16 compares the performances of Lipi Gnani and Tesseract OCRs on the pages of the other language datasets. Plotted in the two subfigures are the Unicode and word error rates obtained by the two OCRs on each of the test pages. It is clear that Lipi Gnani consistently has lower Unicode and word error rates, except for a very few pages.

Table 10 compares the average accuracies of the two OCRs on the three datasets. Though the mean UA for both the OCRs is above 90% on all the datasets, the WA performance of Lipi Gnani is significantly better for Konkani, being 25.54% higher than that of Tesseract. Even for Tulu and Sanskrit, the increase in word accuracy is more than 8%. This shows that a dictionary-based post-processing or a vocabulary-learning classifier such as LSTM is a disadvantage, when dealing with out-of-vocabulary text, which typically happens with text that is transliterated from another language.

## 5 CONCLUSION

A full-fledged, Kannada OCR has been developed, with all the image processing routines implemented using Java. This has resulted in a computationally efficient, readily usable OCR on Windows, Linux and Mac OS. The OCR is invoked inside a user-friendly GUI, also developed by us in Java. The GUI has facility to recognize individual scanned pages or all the pages of an entire book. The latter facility was specifically added to help the NGO's to create Braille versions of school texts for the use of blind children. Thus, the output text of the OCR can be saved in .rtf, .xml or braille format. It also has provision to save the recognized Unicode text, and the line and word boundaries in the industry standard METS/ALTO XML format.

This has led to the use of our Lipi Gnani OCR by several NGO's working for the blind as well as by www.kannadapustaka.org, which has made available all the Kannada school books from eighth standard to PUC as both Braille books and audio books. This contribution was also recognized by the Digital Empowerment Foundation in giving us the Manthan Award in 2014 under the category of E-Inclusion & Accessibility [Ramakrishnan and Shiva Kumar 2014]. Recently, the Department of Kannada and Culture, Government of Karnataka has started using our OCR to digitize the rich Kannada literature books, which are out of copyright, to be put on their official site, www.kanaja.in. Together with the PrintToBraille GUI developed by us, it has already been used by some blind students to convert their text books into Unicode text and for reading by a text-to-speech engine.

Table 10. Unicode and word level accuracies of Lipi Gnani and Tesseract OCRs on Konkani (39 pages; N = 43,428; $N_W$ = 6,647), Tulu (10 pages; N = 9,464; $N_W$ = 1,297) and Sanskrit (60 pages; N = 25,196; $N_W$ = 3,606) datasets. All the notations are the same as in Table 8.

| Dataset | OCR | UA (%) | M | S | I | D | $M_W$ | $S_W$ | $I_W$ | $D_W$ | WA (%) |
|---|---|---|---|---|---|---|---|---|---|---|---|
| Konkani | Lipi Gnani | 94.85 | 43,002 | 1,338 | 236 | 662 | 6,517 | 1,015 | 28 | 158 | 81.93 |
| | Tesseract | 91.06 | 43,675 | 1,672 | 1,228 | 981 | 6,254 | 2,260 | 123 | 516 | 56.39 |
| Tulu | Lipi Gnani | 97.01 | 9,343 | 110 | 26 | 147 | 1,292 | 163 | 3 | 8 | 86.58 |
| | Tesseract | 94.83 | 9,758 | 85 | 349 | 55 | 1,314 | 257 | 22 | 5 | 78.1 |
| Sanskrit | Lipi Gnani | 94.13 | 24,983 | 807 | 230 | 443 | 3,618 | 577 | 32 | 20 | 82.56 |
| | Tesseract | 92.47 | 25,317 | 796 | 611 | 490 | 3,674 | 753 | 120 | 52 | 74.35 |

In future, we intend to introduce an option to invoke the dictionary of a specific language, so that the performance can still be improved, when the document being OCR'ed is strictly from one language. However, this is easier said that done, especially for languages such as Tulu, where there are only a few printed books and Konkani, for which only a subset of books are printed in Kannada script.

## REFERENCES

K. G. Aparna and A. G. Ramakrishnan. 2002. A complete Tamil optical character recognition system. In *Fifth IAPR International Workshop on Document Analysis Systems (DAS 2002)*. Springer-Verlag Berlin, 53–57. https://doi.org/10.1007/3-540-45869-7_6

T.V. Ashwin and P.S. Sastry. 2002. A font and size-independent OCR system for printed Kannada documents using support vector machines. *Sadhana* 27, 1 (Feb 2002), 35–58. https://doi.org/10.1007/BF02703311

Veena Bansal and RMK Sinha. 2000. Integrating knowledge sources in Devanagari text recognition system. *IEEE Trans. on Systems, Man, and Cybernetics-Part A: Systems and Humans* 30, 4 (2000), 500–505.

Veena Bansal and R M K Sinha. 2002. Segmentation of touching and fused Devanagari characters. *Pattern recognition* 35, 4 (2002), 875–893.

B. B. Chaudhuri and U. Pal. 1998. A complete printed Bangla OCR system. *Pattern Recognition* 31, 5 (1998), 531–549. https://doi.org/10.1016/S0031-3203(97)00078-2

B B Chaudhuri, U Pal, and Mandar Mitra. 2002. Automatic recognition of printed Oriya script. *Sadhana* 27, 1 (2002), 23–34.

D Dhanya, A G Ramakrishnan, and Peeta Basa Pati. 2002. Script identification in printed bilingual documents. *Sadhana* 27, 1 (2002), 73–82.

G David Forney. 1973. The Viterbi algorithm. *Proc. IEEE* 61, 3 (1973), 268–278.

Venu Govindaraju and Srirangaraj Ranga Setlur. 2009. *Guide to OCR for Indic Scripts: Document Recognition and Retrieval.* Springer Science & Business Media.

Rangachar Kasturi, Lawrence O'gorman, and Venu Govindaraju. 2002. Document image analysis: A primer. *Sadhana* 27, 1 (2002), 3–22.

Aparna Kokku and Srinivasa Chakravarthy. 2009. A Complete OCR System for Tamil Magazine Documents. In *Guide to OCR for Indic Scripts*. Springer, 147–162. https://doi.org/10.1007/978-1-84800-330-9_7

Gurpreet Singh Lehal and Chandan Singh. 2000. A Gurmukhi script recognition system. In *Proc. 15th International Conf. on Pattern Recognition (ICPR)*, Vol. 2. IEEE, 557–560. https://doi.org/10.1109/ICPR.2000.906135

Gurpreet S Lehal and Chandan Singh. 2002. A post-processor for Gurmukhi OCR. *Sadhana* 27, 1 (2002), 99–111.

Minesh Mathew, Ajeet Kumar Singh, and C. V. Jawahar. 2016. Multilingual OCR for Indic scripts. In *12th IAPR Workshop on Document Analysis Systems (DAS 2016)*. IEEE, 186–191.

IISc MILE. 2018a. Dataset of Konkani documents printed using Kannada script. Retrieved December 30, 2018 from https://github.com/MILE-IISc/KonkaniDocumentsInKannadaScript

IISc MILE. 2018b. Dataset of scanned images of Sanskrit text printed using Kannada script. Retrieved December 30, 2018 from https://github.com/MILE-IISc/SanskritPagesUsingKannadaScript

IISc MILE. 2018c. Dataset of scanned pages of Tulu books. Retrieved December 30, 2018 from https://github.com/MILE-IISc/TuluDocuments

P Nagabhushan and Radhika M Pai. 1999. Modified region decomposition method and optimal depth decision tree in the recognition of non-uniform sized characters–An experimentation with Kannada characters. *Pattern Recognition Letters* 20, 14 (1999), 1467–1475.

Premkumar S Natarajan, Ehry MacRostie, and Michael Decerbo. 2005. The BBN byblos hindi OCR system. In *Document Recognition and Retrieval XII*, Vol. 5676. International Society for Optics and Photonics, 10–17.

N. V. Neeba, Anoop Namboodiri, C. V. Jawahar, and P. J. Narayanan. 2009. Recognition of Malayalam Documents. In *Guide to OCR for Indic Scripts*. Springer, 125–146. https://doi.org/10.1007/978-1-84800-330-9_6

Atul Negi, Chakravarthy Bhagvati, and B Krishna. 2001. An OCR system for Telugu. In *Document Analysis and Recognition, Proc. Sixth International Conf. on*. IEEE, 1110–1114.

Nobuyuki Otsu. 1979. A threshold selection method from gray-level histograms. *IEEE Transactions on Systems, Man, and Cybernetics* 9, 1 (1979), 62–66.

A.G. Ramakrishnan and H.R. Shiva Kumar. 2014. Manthan Award 2014. *E-Inclusion & Accessibilty - Winner* (2014). http://manthanaward.org/e-inclusion-accessibilty-winner-2014/

Tony M Rath and Rudrapatna Manmatha. 2007. Word spotting for historical documents. *International Journal of Document Analysis and Recognition (IJDAR)* 9, 2-4 (2007), 139–152.

Ray Smith. 2016a. Building a Multilingual OCR Engine. Retrieved June 20, 2016 from https://github.com/tesseract-ocr/docs/blob/master/das_tutorial2016/7Building%20a%20Multi-Lingual%20OCR%20Engine.pdf

Ray Smith. 2016b. Tesseract blends old and new OCR technology. *Tutorial at DAS* (2016).

Sargur N Srihari. 1993. Recognition of handwritten and machine-printed text for postal address interpretation. *Pattern recognition letters* 14, 4 (1993), 291–302.

Tesseract. 2018. Tesseract Manual. Retrieved December 15, 2018 from https://github.com/tesseract-ocr/tesseract/blob/master/doc/tesseract.1.asc

Consortium Unicode. 2018. The Unicode Standard v11.0 U0C80. Retrieved June 5, 2018 from https://unicode.org/charts/PDF/U0C80.pdf

B. Vijay Kumar and A.G. Ramakrishnan. 2002. Machine recognition of printed Kannada text. In *Fifth IAPR International Workshop on Document Analysis Systems (DAS-2002)*, Vol. 5. Springer Verlag, Berlin, 37 – 48. https://doi.org/10.1007/3-540-45869-7_4

B. Vijay Kumar and A.G. Ramakrishnan. 2004. Radial basis function and subspace approach for printed Kannada text recognition. In *International Conf. on Acoustics, Speech, and Signal Processing (ICASSP)*, Vol. 5. IEEE, 321 – 324. https://doi.org/10.1109/ICASSP.2004.1327112

Nick White. 2014. uzn format. Retrieved August 21, 2014 from https://github.com/OpenGreekAndLatin/greek-dev/wiki/uzn-format

Wikipedia. 2018a. Konkani language. Retrieved October 3, 2018 from https://en.wikipedia.org/wiki/Konkani_language

Wikipedia. 2018b. Tulu language. Retrieved October 3, 2018 from https://en.wikipedia.org/wiki/Tulu_language

## A KANNADA RECOGNITION UNITS

Table 11. Kannada recognition units: independent vowels, consonants with/without vowel modifiers, pure consonants, ottus.

| ಅ<br>a | ಆ<br>aa | ಇ<br>i | ಈ<br>ii | ಉ<br>u | ಊ<br>uu | ಋ<br>Ru | ಎ<br>e | ಏ<br>ee | ಐ<br>ai | ಒ<br>o | ಓ<br>oo | ಔ<br>au | -a | _ottu |
|---|---|---|---|---|---|---|---|---|---|---|---|---|---|---|
| ಕ /ka/ | ಕಾ | ಕಿ | ಕೀ | ಕು | ಕೂ | ಕೃ | ಕೆ | ಕೇ | ಕೈ | ಕೊ | ಕೋ | ಕೌ | ಕ್ | ್ಕ |
| ಖ /kha/ | ಖಾ | ಖಿ | | ಖು | ಖೂ | | ಖೆ | | | ಖೊ | | ಖೌ | ಖ್ | ್ಖ |
| ಗ /ga/ | ಗಾ | ಗಿ | | ಗು | ಗೂ | | ಗೆ | | | ಗೊ | | ಗೌ | ಗ್ | ್ಗ |
| ಘ /gha/ | ಘಾ | ಘಿ | | ಘು | ಘೂ | | ಘೆ | | | ಘೊ | | ಘೌ | ಘ್ | ್ಘ |
| ಙ /nga/ | ಙಾ | ಙಿ | | ಙು | ಙೂ | | ಙೆ | | | ಙೊ | | ಙೌ | ಙ್ | ್ಙ |
| ಚ /ca/ | ಚಾ | ಚಿ | | ಚು | ಚೂ | | ಚೆ | | | ಚೊ | | ಚೌ | ಚ್ | ್ಚ |
| ಛ /cha/ | ಛಾ | ಛಿ | | ಛು | ಛೂ | | ಛೆ | | | ಛೊ | | ಛೌ | ಛ್ | ್ಛ |
| ಜ /ja/ | ಜಾ | ಜಿ | | ಜು | ಜೂ | | ಜೆ | | | ಜೊ | | ಜೌ | ಜ್ | ್ಜ |
| ಝ /jha/ | ಝಾ | ಝಿ | | ಝು | ಝೂ | | ಝೆ | | | ಝೊ | | ಝೌ | ಝ್ | ್ಝ |
| ಞ /nya/ | ಞಾ | ಞಿ | | ಞು | ಞೂ | | ಞೆ | | | ಞೊ | | ಞೌ | ಞ್ | ್ಞ |
| ಟ /tta/ | ಟಾ | ಟಿ | | ಟು | ಟೂ | | ಟೆ | | | ಟೊ | | ಟೌ | ಟ್ | ್ಟ |
| ಠ /ttha/ | ಠಾ | ಠಿ | | ಠು | ಠೂ | | ಠೆ | | | ಠೊ | | ಠೌ | ಠ್ | ್ಠ |
| ಡ /dda/ | ಡಾ | ಡಿ | | ಡು | ಡೂ | | ಡೆ | | | ಡೊ | | ಡೌ | ಡ್ | ್ಡ |
| ಢ /ddha/ | ಢಾ | ಢಿ | | ಢು | ಢೂ | | ಢೆ | | | ಢೊ | | ಢೌ | ಢ್ | ್ಢ |
| ಣ /nna/ | ಣಾ | ಣಿ | | ಣು | ಣೂ | | ಣೆ | | | ಣೊ | | ಣೌ | ಣ್ | ್ಣ |
| ತ /ta/ | ತಾ | ತಿ | | ತು | ತೂ | | ತೆ | | | ತೊ | | ತೌ | ತ್ | ್ತ |
| ಥ /tha/ | ಥಾ | ಥಿ | | ಥು | ಥೂ | | ಥೆ | | | ಥೊ | | ಥೌ | ಥ್ | ್ಥ |
| ದ /da/ | ದಾ | ದಿ | | ದು | ದೂ | | ದೆ | | | ದೊ | | ದೌ | ದ್ | ್ದ |
| ಧ /dha/ | ಧಾ | ಧಿ | | ಧು | ಧೂ | | ಧೆ | | | ಧೊ | | ಧೌ | ಧ್ | ್ಧ |
| ನ /na/ | ನಾ | ನಿ | | ನು | ನೂ | | ನೆ | | | ನೊ | | ನೌ | ನ್ | ್ನ |
| ಪ /pa/ | ಪಾ | ಪಿ | | ಪು | ಪೂ | | ಪೆ | | | ಪೊ | | ಪೌ | ಪ್ | ್ಪ |
| ಫ /pha/ | ಫಾ | ಫಿ | | ಫು | ಫೂ | | ಫೆ | | | ಫೊ | | ಫೌ | ಫ್ | ್ಫ |
| ಬ /ba/ | ಬಾ | ಬಿ | | ಬು | ಬೂ | | ಬೆ | | | ಬೊ | | ಬೌ | ಬ್ | ್ಬ |
| ಭ /bha/ | ಭಾ | ಭಿ | | ಭು | ಭೂ | | ಭೆ | | | ಭೊ | | ಭೌ | ಭ್ | ್ಭ |
| ಮ /ma/ | ಮಾ | ಮಿ | | ಮು | ಮೂ | | ಮೆ | | | ಮೊ | | ಮೌ | ಮ್ | ್ಮ |
| ಯ /ya/ | ಯಾ | ಯಿ | | ಯು | ಯೂ | | ಯೆ | | | ಯೊ | | ಯೌ | ಯ್ | ್ಯ |
| ರ /ra/ | ರಾ | ರಿ | | ರು | ರೂ | | ರೆ | | | ರೊ | | ರೌ | ರ್ | ್ರ |
| ಱ /rra/ | ಱಾ | ಱಿ | | ಱು | ಱೂ | | ಱೆ | | | ಱೊ | | ಱೌ | ಱ್ | ್ಱ |
| ಲ /la/ | ಲಾ | ಲಿ | | ಲು | ಲೂ | | ಲೆ | | | ಲೊ | | ಲೌ | ಲ್ | ್ಲ |
| ಳ /lla/ | ಳಾ | ಳಿ | | ಳು | ಳೂ | | ಳೆ | | | ಳೊ | | ಳೌ | ಳ್ | ್ಳ |
| ೞ /llla/ | ೞಾ | ೞಿ | | ೞು | ೞೂ | | ೞೆ | | | ೞೊ | | ೞೌ | ೞ್ | ್ೞ |
| ವ /va/ | ವಾ | ವಿ | | ವು | ವೂ | | ವೆ | | | ವೊ | | ವೌ | ವ್ | ್ವ |
| ಶ /sha/ | ಶಾ | ಶಿ | | ಶು | ಶೂ | | ಶೆ | | | ಶೊ | | ಶೌ | ಶ್ | ್ಶ |
| ಷ /ssa/ | ಷಾ | ಷಿ | | ಷು | ಷೂ | | ಷೆ | | | ಷೊ | | ಷೌ | ಷ್ | ್ಷ |
| ಸ /sa/ | ಸಾ | ಸಿ | | ಸು | ಸೂ | | ಸೆ | | | ಸೊ | | ಸೌ | ಸ್ | ್ಸ |
| ಹ /ha/ | ಹಾ | ಹಿ | | ಹು | ಹೂ | | ಹೆ | | | ಹೊ | | ಹೌ | ಹ್ | ್ಹ |

| ಃ | ಂ | ೧ / 1 | ೨ / 2 | ೩ / 3 | ೪ / 4 | ೫ / 5 | ೬ / 6 | ೭ / 7 | ೮ / 8 | ೯ / 9 | ಽ | । |
|---|---|---|---|---|---|---|---|---|---|---|---|---|
| visarga | anusvara/zero | one | two | three | four | five | six | seven | eight | ರ್ / nine | avagraha | danda |

Table 12. Additional Kannada recognition units arising due to the nature of printing in some old letterpress printed documents. The glyphs in the first column and the title row are additional symbols that are used to form the aksharas shown in the rest of the rows and columns.

| | ಾ /aa/ | ೌ /au/ | ್ /halant/ | | ು /u/ | ೂ /uu/ | | ೂ /uu/ |
|---|---|---|---|---|---|---|---|---|
| ಕ /k/ | ಕಾ /kaa/ | ಕೌ /kau/ | ಕ್ /k/ | ಕ /ka/ | ಕು /ku/ | ಕೂ /kuu/ | ಕೆ /ke/ | ಕೊ /ko/ |
| ಖ /kh/ | ಖಾ | ಖೌ | ಖ್ | ಖ | ಖು | ಖೂ | ಖೆ | ಖೊ |
| ಗ /g/ | ಗಾ | ಗೌ | ಗ್ | ಗ | ಗು | ಗೂ | ಗೆ | ಗೊ |
| ಘ /gh/ | ಘಾ | ಘೌ | ಘ್ | ಘ | ಘು | ಘೂ | ಘೆ | ಘೊ |
| ಙ /ng/ | ಙಾ | ಙೌ | ಙ್ | ಙ | ಙು | ಙೂ | ಙೆ | ಙೊ |
| ಚ /c/ | ಚಾ | ಚೌ | ಚ್ | ಚ | ಚು | ಚೂ | ಚೆ | ಚೊ |
| ಛ /ch/ | ಛಾ | ಛೌ | ಛ್ | ಛ | ಛು | ಛೂ | ಛೆ | ಛೊ |
| ಜ /j/ | ಜಾ | ಜೌ | ಜ್ | ಜ | ಜು | ಜೂ | ಜೆ | ಜೊ |
| ಞ /ny/ | ಞಾ | ಞೌ | ಞ್ | ಞ | ಞು | ಞೂ | ಞೆ | ಞೊ |
| ಟ /tt/ | ಟಾ | ಟೌ | ಟ್ | ಟ | ಟು | ಟೂ | ಟೆ | ಟೊ |
| ಠ /tth/ | ಠಾ | ಠೌ | ಠ್ | ಠ | ಠು | ಠೂ | ಠೆ | ಠೊ |
| ಡ /dd/ | ಡಾ | ಡೌ | ಡ್ | ಡ | ಡು | ಡೂ | ಡೆ | ಡೊ |
| ಢ /ddh/ | ಢಾ | ಢೌ | ಢ್ | ಢ | ಢು | ಢೂ | ಢೆ | ಢೊ |
| ಣ /nn/ | ಣಾ | ಣೌ | ಣ್ | ಣ | ಣು | ಣೂ | ಣೆ | ಣೊ |
| ತ /t/ | ತಾ | ತೌ | ತ್ | ತ | ತು | ತೂ | ತೆ | ತೊ |
| ಥ /th/ | ಥಾ | ಥೌ | ಥ್ | ಥ | ಥು | ಥೂ | ಥೆ | ಥೊ |
| ದ /d/ | ದಾ | ದೌ | ದ್ | ದ | ದು | ದೂ | ದೆ | ದೊ |
| ಧ /dh/ | ಧಾ | ಧೌ | ಧ್ | ಧ | ಧು | ಧೂ | ಧೆ | ಧೊ |
| ನ /n/ | ನಾ | ನೌ | ನ್ | ನ | ನು | ನೂ | ನೆ | ನೊ |
| ಪ /p/ | ಪಾ | ಪೌ | ಪ್ | - | - | - | - | - |
| ಫ /ph/ | ಫಾ | ಫೌ | ಫ್ | - | - | - | - | - |
| ಬ /b/ | ಬಾ | ಬೌ | ಬ್ | ಬ | ಬು | ಬೂ | ಬೆ | ಬೊ |
| ಭ /bh/ | ಭಾ | ಭೌ | ಭ್ | ಭ | ಭು | ಭೂ | ಭೆ | ಭೊ |
| ರ /r/ | ರಾ | ರೌ | ರ್ | ರ | ರು | ರೂ | ರೆ | ರೊ |
| ಲ /l/ | ಲಾ | ಲೌ | ಲ್ | ಲ | ಲು | ಲೂ | ಲೆ | ಲೊ |
| ಳ /ll/ | ಳಾ | ಳೌ | ಳ್ | ಳ | ಳು | ಳೂ | ಳೆ | ಳೊ |
| ವ /v/ | ವಾ | ವೌ | ವ್ | - | - | - | - | - |
| ಶ | ಶಾ | ಶೌ | ಶ್ | ಶ | ಶು | ಶೂ | ಶೆ | ಶೊ |
| ಷ /ss/ | ಷಾ | ಷೌ | ಷ್ | ಷ | ಷು | ಷೂ | ಷೆ | ಷೊ |
| ಸ /s/ | ಸಾ | ಸೌ | ಸ್ | ಸ | ಸು | ಸೂ | ಸೆ | ಸೊ |
| ಹ /h/ | ಹಾ | ಹೌ | ಹ್ | ಹ | ಹು | ಹೂ | ಹೆ | ಹೊ |

| | ು /u/ | ಾ /aa*/ | ು /u*/ | ೂ /uu*/ | ೌ /au*/ | ್ /halant*/ |
|---|---|---|---|---|---|---|
| ಝ /jha_part/ | ಝ /jha/ | ಝಾ /jhaa/ | ಝು /jhu/ | ಝೂ /jhuu/ | ಝೌ /jhau/ | ಝ್ /jh/ |
| ವ /va/ | ಮ /ma/ | ಮಾ /maa/ | ಮು /mu/ | ಮೂ /muu/ | ಮೌ /mau/ | ಮ್ /m/ |
| ಯ /ya_part/ | ಯ /ya/ | ಯಾ /yaa/ | ಯು /yu/ | ಯೂ /yuu/ | ಯೌ /yau/ | ಯ್ /y/ |

| | ು /u/ |
|---|---|
| ಝಿ /jhi_part/ | ಝಿ /jhi/ |
| ವಿ /vi/ | ಮಿ /mi/ |
| ಯಿ /yi_part/ | ಯಿ /yi/ |

| | ು /u/ | ೂ /uu/ |
|---|---|---|
| ಝೆ /jhe_part/ | ಝೆ /jhe/ | ಝೊ /jho/ |
| ವೆ /ve/ | ಮೆ /me/ | ಮೊ /mo/ |
| ಯೆ /ye_part/ | ಯೆ /ye/ | ಯೊ /yo/ |

| | ು /u/ | ೂ /uu/ |
|---|---|---|
| ಝ /jha/ | ಝು /jhu/ | ಝೂ /jhuu/ |
| ಮ /ma/ | ಮು /mu/ | ಮೂ /muu/ |
| ಯ /ya/ | ಯು /yu/ | ಯೂ /yuu/ |